\documentclass[letterpaper, 10 pt, conference]{ieeeconf}  % Comment this line out if you need a4paper

\IEEEoverridecommandlockouts                              % This command is only needed if 
\usepackage{makecell}
\usepackage{booktabs}
\usepackage{multirow}
\usepackage{hyperref}
\usepackage{url}
\usepackage{algorithm}
\usepackage{algorithmic}
\usepackage{epsfig}
\usepackage{graphicx}
\usepackage{amsmath}
\usepackage{amssymb}
\usepackage{algorithmic}
\usepackage[switch]{lineno}
\usepackage{subcaption}
\usepackage{makecell, multirow, tabularx}
\usepackage{booktabs}
\usepackage{cite}
\usepackage{color}
\usepackage{amsmath}
\usepackage{array}
\usepackage{xcolor}
\usepackage{xcolor}
\definecolor{darkgreen}{RGB}{0,100,0}
\definecolor{darkred}{RGB}{139,0,0}
\usepackage{graphicx}
\usepackage{amsmath}
\usepackage{cleveref}
\crefname{section}{Sec.}{Secs.}

\newcommand{\eg}{e.g.}
\newcommand{\ie}{i.e.}

\Crefname{section}{Section}{Sections}
\Crefname{table}{Table}{Tables}
\crefname{table}{Tab.}{Tabs.}
\crefname{figure}{Fig.}{Figs.}
\crefname{algorithm}{Alg.}{Algs.}
\crefformat{equation}{Eq.~#2#1#3}
\usepackage[flushleft]{threeparttable}
\definecolor{hollywoodcerise}{rgb}{0.96, 0.0, 0.63}
\definecolor{lasallegreen}{rgb}{0.03, 0.47, 0.19}
\definecolor{hanpurple}{rgb}{0.32, 0.09, 0.98}
\definecolor{green(pigment)}{rgb}{0.0, 0.65, 0.31}
\usepackage{amssymb}% http://ctan.org/pkg/amssymb
\usepackage{pifont}% http://ctan.org/pkg/pifont
\newcommand{\cmark}{\ding{51}}%
\newcommand{\xmark}{\ding{55}}%

\title{\LARGE \bf
Energy-based Domain-Adaptive Segmentation with Depth Guidance}

 \author{Jinjing Zhu, Zhedong Hu, Tae-Kyun Kim, and Lin Wang* % <-this % stops a space
 \thanks{*Corresponding author.}% <-this % stops a space
  \thanks{J. Zhu is with HKUST(GZ), China (zhujinjing.hkust@gmail.com)}
  \thanks{Z. Hu is with NCEPU, China (caughyhzd@foxmail.com)}
  \thanks{T. Kim is with KAIST, Korea and ICL, UK (kimtaekyun@kaist.ac.kr)} 
 \thanks{L. Wang is with AI/CMA Thrust, HKUST(GZ), Guangzhou and  HKUST, HongKong, China (linwang@ust.hk)}
 \thanks{This work was done when Z. Hu was interning at VLIS, HKUST(GZ).}
 }

\begin{document}

% \twocolumn[{
% \renewcommand\twocolumn[1][]{#1}%
% \maketitle
% \captionsetup{font=small}
% \begin{center}
% \vspace{-30pt}
%     \centering
%     \includegraphics[ width=0.8\textwidth]{Coverfigure.pdf}
%     \vspace{-12pt}
%     \captionof{figure}{(a) Previous works, \eg, DADA\cite{VuJBCP19}, ignore the discrepancy between semantic and depth features, as well as the reliability of feature fusion with depth guidance. (b) To address this issue, Energy-Based Feature Fusion (EB2F) is proposed to decrease the discrepancy between semantic and depth features, and Reliable Fusion Assessment (RFA) enables the fusion to facilitate segmentation.}
% \label{fig:Introduction}
% \end{center}
% }]

\maketitle

\thispagestyle{empty}
\pagestyle{empty}

\begin{abstract}

Recent endeavors have been made to leverage self-supervised depth estimation as guidance in unsupervised domain adaptation (UDA) for semantic segmentation. Prior arts, however, overlook the discrepancy between semantic and depth features, as well as the reliability of feature fusion, thus leading to suboptimal segmentation performance. To address this issue, we propose a novel UDA framework called \textbf{SMART} (cro\textbf{S}s do\textbf{M}ain sem\textbf{A}ntic segmentation based on ene\textbf{R}gy es\textbf{T}imation) that utilizes Energy-Based Models (EBMs) to obtain task-adaptive features and achieve reliable feature fusion for semantic segmentation with self-supervised depth estimates. Our framework incorporates two novel components: energy-based feature fusion (\textbf{EB2F}) and energy-based reliable fusion Assessment (\textbf{RFA}) modules. The EB2F module produces task-adaptive semantic and depth features by explicitly measuring and reducing their discrepancy using Hopfield energy for better feature fusion. The RFA module evaluates the reliability of the feature fusion using an energy score to improve the effectiveness of depth guidance. Extensive experiments on two datasets demonstrate that our method achieves significant performance gains over prior works, validating the effectiveness of our energy-based learning approach.
\end{abstract}


\vspace{-5pt}
\section{Introduction}
\label{sec:intro}
Semantic segmentation \cite{Chen2016DeepLabSI, zhu2023good} is a crucial yet challenging computer vision task that is vital for robotics and self-driving systems, as it ensures accurate perception for improving safety and efficiency. Deep neural network (DNN)-based methods have been successfully applied thanks to a large amount of high-quality labeled data~\cite{ChenPKMY18}. However, in real-world scenarios, the performance of these methods is often hindered by domain shifts, such as illumination variances or scene differences, between the training and test data.  Unsupervised domain adaptation (UDA) \cite{PanY10, zhu2023patch} has emerged as a solution to mitigate the domain shift issue between the labeled source and unlabeled target domains. Recently, the supplementary information, \eg, depth, has been exploited to bring the geometric cues to enhance the performance of semantic segmentation and facilitate domain adaptation. Since the tasks of semantic segmentation and depth estimation are closely related and complementary, it is beneficial to consider them jointly for scene understanding.

Representative works \cite{SahaOPKCGG21, VuJBCP19} have leveraged supervised depth information from the synthetic domain due to unavailable depth labels in the real domain. However, in reality, depth ground-truth (GT) is not always available for both domains. To address this limitation, some studies have proposed to leverage self-supervised depth estimation approaches \cite{CardaceLRSS22, 0013DHGF21} in the absence of GT depth labels. 

\begin{figure}[t]
    \centering
    \captionsetup{font=small}
    \includegraphics[width=0.99\linewidth]{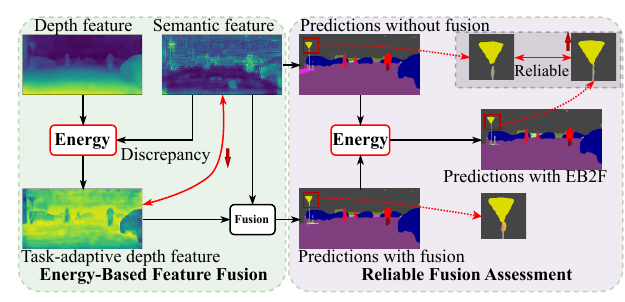}
    \vspace{-17pt}
    \caption{Energy-Based Feature Fusion (EB2F) is proposed to decrease the discrepancy between semantic and depth features, and Reliable Fusion Assessment (RFA) enables the fusion to facilitate segmentation.}
    \label{fig:Introduction}
    \vspace{-20pt}
\end{figure}
% For example, Wang~\cite{0013DHGF21} utilized self-supervised depth guidance to approximate the task feature correlation and adaptation difficulty.

One of the primary problems in utilizing depth information for semantic segmentation is ignoring the discrepancy between depth and semantic features, as they inherently extract semantic and geometric information respectively \cite{WangSLCPY15}. Although existing methods, such as \cite{SahaOPKCGG21, VuJBCP19, 0013DHGF21}, incorporate fusion modules to integrate depth guidance into the segmentation network, they often fail to fully consider the discrepancy between depth and semantic features, resulting in suboptimal segmentation performance. In CTRL \cite{SahaOPKCGG21}, the segmentation accuracy even degrades with the fusion of depth and segmentation features. Furthermore, previous works have not explicitly evaluated the reliability of feature fusion, which hinders our understanding of the role of fusion in improving segmentation. Predictions of some regions with feature fusion are worse than predictions without feature fusion. Intuitively, we ask a question: \textit{`how to explicitly 1) learn task-adaptive features and 2) assess the quality of fusion in domain adaptive semantic segmentation with self-supervised depth estimation?’} 

We advocate a \textit{unified} and \textit{effective} framework for achieving these objectives by employing Energy-Based Models (EBMs) \cite{lecun2006tutorial}. In principle, EBMs aim to quantify the compatibility between variables by building an energy function and give the lowest energy to correct answers and higher energy values to other incorrect answers. Specifically, the Hopfield network \cite{hopfield1982neural,RamsauerSLSWGHA21} employs an energy function to assess the closeness between input patterns and stored patterns and designs an update rule to modify the input patterns to align with the stored patterns. In addition, free energy has been shown to be more effective in detecting out-of-distribution data than softmax scores \cite{LiuWOL20}. The aforementioned two objectives can be subtly achieved via a unified EBM framework enhancing segmentation with depth guidance.

To this end, we propose a novel UDA framework called SMART (cro\textbf{S}s do\textbf{M}ain sem\textbf{A}ntic segmentation based on ene\textbf{R}gy es\textbf{T}imzation) that utilizes EBMs to extract task-adaptive features and perform reliable fusion, facilitating segmentation. SMART comprises two technical novelties that enhance the performance of UDA for semantic segmentation (See Fig. \ref{fig:Introduction}). Firstly, we leverage the Hopfield energy approach \cite{RamsauerSLSWGHA21} to explicitly measure the discrepancy between semantic and depth features, allowing for extracting the task-adaptive semantic and depth features. Additionally, we introduce an energy-based feature fusion (\textbf{EB2F}) to extract task-adaptive features for improved fusion. Secondly, we utilize an energy score to evaluate the reliability of feature fusion by comparing the predictions made with and without feature fusion. We propose an energy-based reliable fusion assessment (\textbf{RFA}) module that transfers reliable guidance to other tasks, thus improving performance using the fusion.

Our main contributions are four-fold: (\textbf{I}) We introduce a novel energy-based framework to solve the cross-domain adaptation problem for semantic segmentation with self-supervised depth estimation. (\textbf{II}) We propose the Energy-Based feature fusion (EB2F) module to facilitate the feature fusion with task-adaptive semantic and depth features. (\textbf{III}) We introduce the Energy-Based reliable fusion assessment (RFA) module to evaluate the reliability of fusion by comparing the predictions made with and without the feature fusion. (\textbf{IV}) Our proposed method consistently achieves state-of-the-art performance on two benchmark datasets.

\section{Related Work}
\label{sec:Related work}

\noindent{\bf UDA for Semantic Segmentation.} 
% \subsection{ UDA for Semantic Segmentation}
Numerous methods~\cite{wang2022bilateral, chen2023ida} have been proposed to address this task. One line of research is to align the pixels of input images between the source and target domains by utilizing generative networks \cite{hoffman2018cycada,zheng2023both}. Other approaches explore adversarial learning to reduce the domain gap~\cite{isobe2021multi, vu2019advent,tsai2018learning}, which trains a discriminator to discern the domain (\ie, source or target) from the data. In particular, in \cite{chen2018road}, the alignment is done in the feature space, while in~\cite{vu2019advent} the alignment is achieved with the output space. In addition, self-training is explored to generate pseudo labels for unlabeled data \cite{zhang2021prototypical}. 

% These approaches improve the model performance by progressively improving the pseudo labels of target data. 

\noindent{\bf Depth-guided UDA for Semantic Segmentation.} 
% \subsection{Depth-guided UDA for Semantic Segmentation}
Depth estimation has been increasingly popular in facilitating domain adaptation for semantic segmentation. SPIGAN \cite{LeeRLG19} and DADA\cite{VuJBCP19} are among the first to use privileged depth information to guide UDA for semantic segmentation.
GIO-Ada \cite{ChenLCG19} and CTRL \cite{SahaOPKCGG21} further exploit adversarial learning to align the domains with depth estimation.
However, due to the lack of supervised depth information in practical applications, recent works leverage self-supervised depth estimation to assist UDA for semantic segmentation. D4 \cite{CardaceLRSS22} and GUDA \cite{CardaceLRSS22} exploit strong depth priors based on self-supervised learning. 
Recently, CorDA \cite{0013DHGF21} applies the depth prediction discrepancy to measure the pixel-wise adaptation difficulty for refining the target pseudo labels. Unfortunately, previous works focuses on leveraging depth information for semantic segmentation while disregarding the discrepancy between depth and semantic features and the reliability of depth guidance. \textit{As a complement to these methods, our work investigates the potential of energy-based learning for improving feature fusion and depth guidance. Our method estimates energy to extract task-adaptive features for the fusion and ensure reliable fusion for better depth guidance.}
\begin{figure*}[t]
    \centering
    \captionsetup{font=small}
    \includegraphics[width=0.8\linewidth]{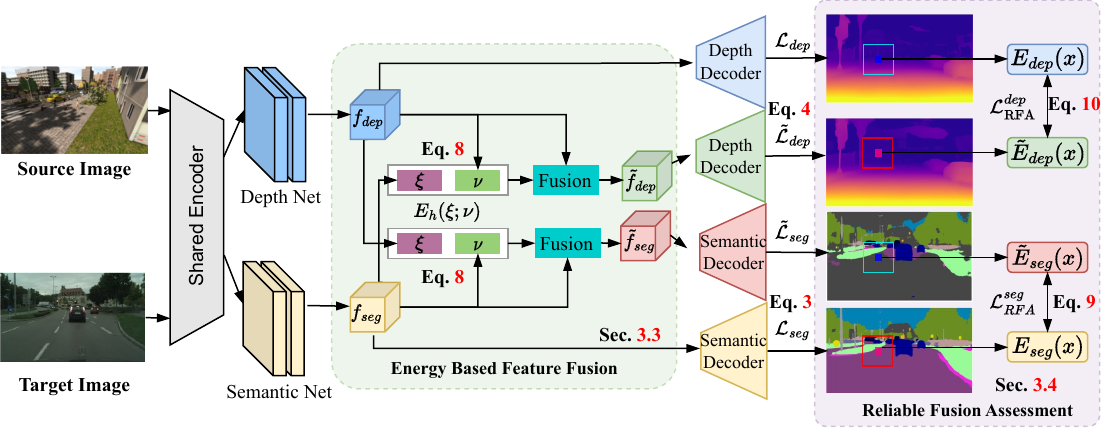}
    \vspace{-5pt}
    \caption{\textbf{Architecture of the proposed SMART framework}, consisting of the shared encoder, task nets, and task decoders.}
    \label{fig:framework}
    \vspace{-17pt}
\end{figure*}

\noindent{\bf Energy-based Learning.} 
%\subsection{Energy-based Learning}
It aims to build a function that maps each point $\mathrm{x}$ of an input space to a single scalar called the \emph{energy}. 
The basic idea of energy-based models (EBMs)~\cite{lecun2006tutorial} is to measure the compatibility between variables (e.g. images and labels) by energy function. 
The energy function output with a lower scalar value indicates a high degree of compatibility. The variables with low compatibility are assigned a high energy value. Recently, it has been widely recognized that EBMs has a significant role to play in generative models \cite{che2020your}. The output of the discriminator is shaped into the energy function to distinguish low-energy real data and high-energy generated data. Moreover, EBMs are also used for object detection tasks~\cite{sun2021react}.
Since energy-based learning can be used to describe the most valuable distribution in a finite target sample, a recent work \cite{xie2022active} attempted to apply it to active domain adaptation and proposed to align domain distributions by reducing free energy biases between domains. Additionally, \cite{LiuWOL20} has provided both mathematical insights and empirical evidence that the energy score is superior to the softmax-based score in detecting out-of-distribution inputs. Furthermore, EBMs are becoming increasingly popular for solving problems in contrastive learning \cite{abs-2202-04933}, continual learning \cite{Joseph0KAB22}, adversarial training tasks \cite{00010YL22}, and domain adaptation \cite{LiDLZL22}. \textit{Differently, we make the \textbf{first} attempt to achieve cross-domain semantic segmentation with depth guidance from an energy-based perspective. Our method significantly enhances the effectiveness of depth guidance with task-adaptive features and reliable fusion based on energy estimation.} 
\vspace{-2pt}
\section{Methodology}
\vspace{-2pt}
\label{sec:meth}
\subsection{Energy-based Models: Preliminaries}
The EBMs aim to build a function $E(\mathbf{x})$ that maps the variable $\mathbf{x}$ to a single non-probabilistic scalar, called the \textit{energy}. A collection of energy values can be turned into a probability density $p(\mathbf{x})$ via Gibbs distribution: 
{\setlength\abovedisplayskip{0pt}
\setlength\belowdisplayskip{0pt}
\begin{equation}
 p(y \mid \mathbf{x})=\frac{e^{-E(\mathbf{x}, y) / \tau}}{\int_{y^{\prime}} e^{-E\left(\mathbf{x}, y^{\prime}\right) / \tau}}=\frac{e^{-E(\mathbf{x}, y) / \tau}}{e^{-E(\mathbf{x}) / \tau}}, 
 \label{Eq1}
\end{equation}}
where the denominator $\int_{y^{\prime}} e^{-E\left(\mathbf{x}, y^{\prime}\right) / \tau}$ is called the partition function, which marginalizes over label $y$, and $\tau$ is the temperature parameter and is set as 1. 
% In this work, we mainly focus on segmentation tasks and get categorical predictions by the neural network $f(x)$, which maps an input x into K classes known as logits. 

% The \textit{Helmholtz free energy} $E(\mathbf{x})$ of a given data point $\mathbf{x}$ can be expressed as the negative of the log partition function:
% {\setlength\abovedisplayskip{2pt}
% \setlength\belowdisplayskip{2pt}
% \begin{equation}
% E(\mathbf{x})=-\log \int_{y^{\prime}} e^{-E\left(\mathbf{x}, y^{\prime}\right)}. 
% \end{equation}}
In this work, we mainly focus on the segmentation task and get per-class predictions $P_{seg}(\mathbf{x})$ by a neural network, which maps an input $\mathbf{x}$ into $K$ classes known as logits. These logits are used to derive a per-class distribution using the softmax function:
{\setlength\abovedisplayskip{4pt}
\setlength\belowdisplayskip{3pt}
\begin{equation}
p(y \mid \mathbf{x})=\frac{e^{P_{seg}^y(\mathbf{x})}}{\sum_i^K e^{P_{seg}^i(\mathbf{x})}}, 
 \label{Eq2}
\end{equation}}
where $P_{seg}^y(\mathbf{x})$ indicates the $y^{\text {th }}$ index of $P_{seg}(\mathbf{x})$, \ie, the logit corresponding to $y^{\text {th }}$ class label.
By connecting Eq.\ref{Eq1} and Eq. \ref{Eq2}, we can define an energy for a given input $(\mathbf{x}, y)$ as $E_{seg}(\mathbf{x}, y)=-P^y_{seg}(\mathbf{x})$.
We design EBMs ~\cite{lecun2006tutorial} to give the lowest energy to the correct answer and higher energy to all other (incorrect) answers by utilizing a negative log-likelihood (NLL) loss, which is defined as
{\setlength\abovedisplayskip{4pt}
\setlength\belowdisplayskip{4pt}
\begin{equation}
\mathcal{L}_{\mathrm{nll}}=E_{seg}(\mathbf{x}, y)+\log \sum_{j=1}^K e^{-E_{seg}(\mathbf{x}, j)}.    \nonumber
\end{equation}}
The negative of the second term can be interpreted as the \textit{free energy} of the ensemble of energies. 

%/vspace{-5pt}
\subsection{Overview}
\label{Overview}
In this work, we are given a labeled synthetic source domain $\mathcal{D}_S=\left\{\left(\mathbf{x}_1^S, \mathbf{y}_1^S, \mathbf{d}_1^S\right), \ldots,\left(\mathbf{x}_n^S, \mathbf{y}_n^S, \mathbf{d}_n^S\right)\right\}$ and an unlabeled target domain $\mathcal{D}_T=\left\{\left(\mathbf{x}_1^T, \mathbf{d}_1^T\right), \ldots,\left(\mathbf{x}_m^T, \mathbf{d}_m^T\right)\right\}$, where $\mathbf{x}_i^S$ ($\mathbf{x}_i^T$) is the $i$-th source (target) sample, $\mathbf{y}_i^S$ is the corresponding source label for semantic segmentation with $K$ classes, $\mathbf{d}_i^S$ ($\mathbf{d}_i^T$) is the $i$-th label for depth estimation, and $n$ ($m$) represents the total number of source (target) samples. Note that precise depth information is often not provided in the real-world target dataset. Therefore, as proposed in CorDA \cite{0013DHGF21}, we utilize self-supervised depth estimation to generate pseudo depth labels on the target domain $\mathbf{d}_i^T$ or the source domain (\eg, GTA5 \cite{richter2016playing}), where the depth information is also unavailable. Additionally, we employ the self-training approach~\cite{ZouYLKW19} to generate target semantic pseudo-labels based on the target semantic predictions.

Our proposed SMART architecture is shown in Fig. \ref{fig:framework}. Our key ideas are two folds. First, to achieve task-adaptive feature fusion, we propose an energy-based feature fusion (EB2F) module to measure the relationship between the semantic and depth features. Second, we propose an energy-based reliable fusion assessment (RFA) loss to evaluate the feature fusion by comparing the energy of the predictions with or without fusion. Below are technical details for our proposed method.
Based on EBMs, the objectives of semantic segmentation for two domains are formulated as:
{\setlength\abovedisplayskip{6pt}
\setlength\belowdisplayskip{4pt}
\begin{equation}
\small
\begin{aligned}
&\mathcal{L}_{seg} = \sum_{h=1}^H \sum_{w=1}^W \left(E_{seg}(\mathbf{x},\mathbf{y})+\log \sum_{j=1}^K e^{-E_{seg}(\mathbf{x},  j)}\right), 
\end{aligned}
\label{eq:seg}
\end{equation}}
where $H$ and $W$ are the height and width of images. Since depth estimation is a regression problem, its energy function is defined differently as:
{\setlength\abovedisplayskip{2pt}
\setlength\belowdisplayskip{2pt}
\begin{equation}
\small
\begin{split}
      \mathcal{L}_{dep} &=\sum_{h=1}^H \sum_{w=1}^W E_{dep}(\mathbf{x},\mathbf{d})\\
    &=\sum_{h=1}^H \sum_{w=1}^W \left(\text{berHu} \left(P_{dep}(\mathbf{x})-\mathbf{d}\right)\right),  
\end{split}
\label{eq:dep}
\end{equation}
where $P_{dep}(\mathbf{x})$ is the depth prediction from a depth decoder, and berHu is the reverse Huber loss for depth:
{\setlength\abovedisplayskip{2pt}
\setlength\belowdisplayskip{2pt}
\begin{equation}
\operatorname{berHu}\left(e_z\right)= \begin{cases}\left|e_z\right|, & \text { if }\left|e_z\right| \leq c \\ \frac{e_z^2+c^2}{2 c} & \text { otherwise }\end{cases}, \nonumber
\end{equation}}
where $c$ is a threshold set to $\frac{1}{5}$ of the maximum depth difference. Then, the objective of semantic segmentation and that of depth estimation are formulated respectively as:
{\setlength\abovedisplayskip{1pt}
\setlength\belowdisplayskip{1pt}
\begin{equation}
\begin{aligned}
\mathcal{L}_{seg}^{total} &= \mathcal{L}_{seg}^{S}+\mathcal{L}_{seg}^{T}+\tilde{\mathcal{L}}_{seg}^{S}+\tilde{\mathcal{L}}_{seg}^{T}. \\
\mathcal{L}_{dep}^{total} &= \mathcal{L}_{dep}^{S}+\mathcal{L}_{dep}^{T}+\tilde{\mathcal{L}}_{dep}^{S}+\tilde{\mathcal{L}}_{dep}^{T}.
\label{eq:seg_dep}
\end{aligned}
\end{equation}}
Note $\mathcal{L}$ ($\tilde{\mathcal{L}}$) is the objective of the framework without (with) fusion, $\mathcal{L}^{S}$($\mathcal{L}^{T}$) denotes for source (target) domains. Overall, through Eq.~\ref{eq:seg_dep}, we deploy the supervised loss to train the proposed SMART framework:
{\setlength\abovedisplayskip{2pt}
\setlength\belowdisplayskip{2pt}
\begin{equation}
\begin{aligned}
\small
\mathcal{L}_{s}=\mathcal{L}_{seg}^{total}+\alpha\mathcal{L}_{dep}^{total}. \nonumber
\end{aligned}
\end{equation}}

% %/vspace{-3pt}
\subsection{Energy-Based Feature Fusion}
\label{EB2F}
In contrast to prior methods that directly fuse features, we notice the discrepancy between semantic and depth features and propose a novel approach to obtain more task-adaptive semantic and depth features before fusion. Our approach draws inspiration from the Hopfield network \cite{hopfield1982neural}, which is originally proposed for pattern storage and retrieval. Specifically, we utilize the Hopfield energy \cite{RamsauerSLSWGHA21} to describe the relationships between semantic and depth representations in our proposed method. This is a crucial step in addressing the issue of discrepant feature fusion, which can significantly impair segmentation performance. In accordance with the definitions of the Hopfield network, $\boldsymbol{\xi}$ and $\boldsymbol{\nu}$ denote the input pattern and stored pattern. The Hopfield energy function $E_{h}(\boldsymbol{\xi}; \boldsymbol{\nu})$ is then employed to measure the similarity between these patterns, ensuring better fusion of task-adaptive semantic and depth features for improved segmentation performance. Our method provides a means to the challenge of feature fusion in UDA for segmentation with depth guidance. %Our proposed method has the potential to enhance the accuracy and reliability of segmentation results, making it a valuable contribution to the field.
The Hopfield energy function $E_{h}(\boldsymbol{\xi}; \boldsymbol{\nu})$ is computed using the log-sum-exp (lse) function as follows:
{\setlength\abovedisplayskip{1pt}
\setlength\belowdisplayskip{1pt}
\begin{equation}
E_{h}(\boldsymbol{\xi}; \boldsymbol{\nu})=\frac{1}{2} \boldsymbol{\xi}^T \boldsymbol{\xi}-\operatorname{lse}\left(\boldsymbol{\nu}^T \boldsymbol{\xi}\right),
\label{eq_en}
\end{equation}}
To make the patterns closer, we update the input pattern by taking the derivative $\nabla_{\boldsymbol{\xi}} E_h(\boldsymbol{\xi} ; \boldsymbol{\nu})$ of the energy function with respect to $\boldsymbol{\xi}$ in Eq. \ref{eq_en}. 
{\setlength\abovedisplayskip{1pt}
\setlength\belowdisplayskip{1pt}
\begin{equation}
\nabla_{\boldsymbol{\xi}} E_h(\boldsymbol{\xi} ; \boldsymbol{\nu})=\boldsymbol{\xi}-\boldsymbol{\nu} \operatorname{softmax}\left(\boldsymbol{\nu}^T \boldsymbol{\xi}\right),
\label{eq11}
\end{equation}}
where $\nabla_{\boldsymbol{\xi}}$ denotes the gradient with respect to $\boldsymbol{\xi}$, and $\operatorname{softmax}$ denotes the softmax function.
We update the input pattern using a gradient descent with a step size $\gamma$:
{\setlength\abovedisplayskip{1pt}
\setlength\belowdisplayskip{1pt}
\begin{equation}
\begin{split}
\boldsymbol{\xi}_{n+1}&=\boldsymbol{\xi}_n-\gamma\left(\boldsymbol{\xi}_n-\boldsymbol{\nu} \operatorname{softmax}\left(\boldsymbol{\nu}^T \boldsymbol{\xi}_n\right)\right)\\
&=(1-\gamma)\boldsymbol{\xi}_n+\gamma\boldsymbol{\nu} \operatorname{softmax}\left(\boldsymbol{\nu}^T \boldsymbol{\xi}_n\right),   
\end{split}
\end{equation}}
where $\boldsymbol{\xi}_{n+1}$ and $\boldsymbol{\xi}_n$ are the updated input pattern after $n+1$ iterations and at the $n$-th iteration respectively.
% After setting $\gamma=1$, we get the updated input pattern: 
% $$
% \boldsymbol{\xi}_{n+1}=\boldsymbol{\nu} \text { softmax }\left(\boldsymbol{\nu}^T \boldsymbol{\xi}_n\right).
% $$
We propose the energy-based feature fusion (EB2F) module based on the updated input patter. The fusion module is utilized to combine the updated input pattern and the stored pattern to obtain an updated stored pattern.
The updated stored pattern is formulated as:
\begin{equation}
\begin{split}
   \tilde{\nu} &= EB2F (\xi, \nu) \\
   &= \textit{Fusion}((1-\gamma)\boldsymbol{\xi}_n+\gamma\boldsymbol{\nu} \operatorname{softmax}\left(\boldsymbol{\nu}^T \boldsymbol{\xi}_n\right), \nu). \nonumber  
\end{split}
\end{equation} 
To further improve the fusion process, this work employs two fusion schemes as follows:

\noindent\textbf{\textit{Scheme 1:}} Directly adding features $\boldsymbol{\xi}$ and $\boldsymbol{\nu}$ is defined as
{\setlength\abovedisplayskip{2pt}
\setlength\belowdisplayskip{2pt}
$$
{\textit{Fusion}_{1}}(\boldsymbol{\xi},\boldsymbol{\nu}) = \boldsymbol{\xi}+\boldsymbol{\nu}.
$$
\noindent\textbf{\textit{Scheme 2:}} Building upon the attention module \cite{0013DHGF21}, we propose to fuse the features $\boldsymbol{\xi}$ and $\boldsymbol{\nu}$ as follows:
{\setlength\abovedisplayskip{2pt}
\setlength\belowdisplayskip{2pt}
\begin{equation}
    \begin{split}
        {\textit{Fusion}_{2}}(\boldsymbol{\xi},\boldsymbol{\nu}) =  \boldsymbol{\nu} + (W_d^1 \otimes \boldsymbol{\xi})\odot \sigma(W_d^2\otimes\boldsymbol{\xi}). \nonumber
    \end{split}
\end{equation}}

\begin{table*}[t!]
  \centering
  \captionsetup{font=small}
  \resizebox{0.99\linewidth}{!}{ 
  \begin{tabular}{l|ccccccccccccccccccc|c}
\toprule
Method      & \rotatebox{90}{road} & \rotatebox{90}{.walk} & \rotatebox{90}{build.} & \rotatebox{90}{wall} & \rotatebox{90}{fence} & \rotatebox{90}{pole} & \rotatebox{90}{light} & \rotatebox{90}{sign} & \rotatebox{90}{veget.} & \rotatebox{90}{terrain} & \rotatebox{90}{sky}  & \rotatebox{90}{person} & \rotatebox{90}{rider} & \rotatebox{90}{car}  & \rotatebox{90}{truck} & \rotatebox{90}{bus}  & \rotatebox{90}{train} & \rotatebox{90}{moto.}& \rotatebox{90}{bike} & mIoU \\ \hline\hline
Source \cite{tsai2018learning}  & 75.8 & 16.8    & 77.2   & 12.5 & 21.0  & 25.5 & 30.1  & 20.1 & 81.3   & 24.6    & 70.3 & 53.8   & 26.4  & 49.9 & 17.2  & 25.9 & 6.5   & 25.3  & 36.0   & 36.6 \\
BDL   \cite{li2019bidirectional}      & 91.0 & 44.7    & 84.2   & 34.6 & 27.6  & 30.2 & 36.0  & 36.0 & 85.0   & 43.6    & 83.0 & 58.6   & 31.6  & 83.3 & 35.3  & 49.7 & 3.3   & 28.8  & 35.6   & 48.5 \\
MRKLD-SP \cite{zou2019confidence}   & 90.8 & 46.0    & 79.9   & 27.4 & 23.3  & \textbf{42.3} & 46.2  & 40.9 & 83.5   & 19.2    & 59.1 & 63.5   & 30.8  & 83.5 & 36.8  & 52.0 & 28.0  & 36.8  &  46.4  & 49.2 \\
Kim et al. \cite{kim2020learning} & 92.9 & 55.0    & 85.3   & 34.2 & 31.1  & 34.9 &  40.7     &  34.0    &  85.2      &  40.1       &  87.1    &   61.0     &   31.1    &  82.5    &  32.3     &  42.9    &   0.3    &   36.4    & 46.1    &50.2 \\
CAG-UDA \cite{zhang2019category}    & 90.4 & 51.6    & 83.8   & 34.2 & 27.8  & 38.4 &   25.3    &    48.4  &     85.4   &   38.2      &   78.1   &    58.6    &   34.6    &    84.7  &   21.9    &   42.7   &  \textbf{  41.1}   &   29.3    &  37.2   &50.2 \\
FDA  \cite{yang2020fda}        & 92.5 & 53.3    & 82.4   & 26.5 & 27.6  & 36.4 &   40.6    &    38.9  &   82.3     &    39.8     &  78.0    &    62.6    &    34.4   &   84.9   &  34.1     &  53.1    &  16.9     &   27.7    &   46.4  & 50.5\\
PIT   \cite{lv2020cross}      &   87.5   &     43.4    &   78.8     &   31.2   &  30.2     &  36.3    &   39.9    &   42.0   &   79.2     &    37.1     &  79.3    &   65.4     &    \textbf{37.5}   &   83.2   &    46.0   &   45.6   &  25.7     &  23.5     &  49.9   & 50.6\\
IAST   \cite{mei2020instance}      &   93.8   &    57.8     &   85.1     &  \textbf{39.5}    &   26.7    &   26.2   &   43.1    &   34.7   &   84.9     &    32.9     &   88.0   &    62.6    &   29.0    &   87.3   &  39.2     &   49.6   &   23.2    &   34.7    &  39.6  & 51.5  \\ 
DACS    \cite{tranheden2021dacs}      &    89.9  &     39.7    &   87.9     &   30.7   &  39.5   &  38.5    &    46.4   &   52.8   &   \textbf{ 88.0 }   &     44.0    &  88.8    & \textbf{  67.2}     &    35.8   &   84.5   &   45.7    &   50.2   &  0.0     &   27.3   & 34.0 & 52.1\\ 
\hline
CorDA    \cite{0013DHGF21} &\textbf{95.2}&\textbf{65.1}&87.4&33.0&40.0&39.9&47.7&\textbf{53.6}&87.6&45.3&88.7&66.2&33.5&89.9&53.9&53.3&0.2&41.0&\textbf{55.8}&56.7\\
SMART+${\textit{Fusion}_{1}}$& 94.3&59.7&\textbf{88.4}&37.9&\textbf{42.9}&40.0&48.4&51.5&87.3&\textbf{48.1}&88.8&66.6&28.8&\textbf{91.5}&\textbf{67.4}&\textbf{57.9}&0.0&\textbf{44.3}&51.9&\textbf{57.7}\\
SMART+${\textit{Fusion}_{2}}$& 94.5&60.3&88.3&31.8&42.8&41.0&\textbf{48.5}&50.5&87.3&47.3&\textbf{89.0}&66.4&28.0&91.4&63.9&56.1&0.0&44.0&51.2
&57.0
\\
\bottomrule

\end{tabular}}
\vspace{-5pt}
  \caption{Semantic segmentation results on the GTA5-to-Cityscapes.}
\vspace{-5pt}
  \label{tab:gta5}
\end{table*}

Here, $\otimes$ denotes the convolution operation, and $\odot$ denotes the element-wise multiplication. $\sigma$ refers to the sigmoid function used for the normalization of attention maps. $W_{d}^1$ and $W_{d}^2$ represent the learnable convolution weights.
% \noindent\textbf{Scheme 2:} design cross attention module to fuse these two features $\xi$ and $\nu$ as
% {\setlength\abovedisplayskip{2pt}
% \setlength\belowdisplayskip{2pt}
% $$
% {\textit{Fusion}_{2}}(\xi,\nu) = \nu + \operatorname{softmax}\left(\frac{\nu {\xi}^T}{\sqrt{d_k}}\right) \xi
% $$
% }
\begin{figure}[t]
    \centering
     \captionsetup{font=small}
    \includegraphics[width=0.7\linewidth]{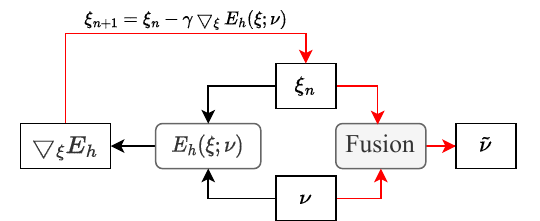}
    \vspace{-5pt}
    \caption{The illustration of the proposed EB2F module.}
    \label{fig:fusion}
    \vspace{-15pt}
\end{figure}
Based on the above EB2F, we obtain the fused semantic and depth features as: 
{\setlength\abovedisplayskip{2pt}
\setlength\belowdisplayskip{2pt}
\begin{equation}
\begin{split}
\small
\tilde{f}_{seg} = \text{EB2F}(f_{dep},f_{seg}), \tilde{f}_{dep} = \text{EB2F}(f_{seg},f_{dep}), \nonumber
\end{split}
\end{equation}}
where ${f}_{seg}$ and ${f}_{dep}$ represent the semantic and depth features extracted by the semantics and depth networks, respectively, and $\tilde{f}_{seg}$ and $\tilde{f}_{dep}$ are the fused features for the segmentation and depth tasks, respectively.
\subsection{Reliable Fusion Assessment}
\label{RFA}
Due to the disparity between semantic and depth features, the feature fusion might slightly improve or even degrade segmentation performance.  Therefore, we aim to measure the reliability of the feature fusion for facilitating segmentation. However, there is a challenging problem:\textit{`how to explicitly measure the reliability of feature fusion for segmentation?'}. To this end, we subtly compare the energy of predictions made with and without fusion. The reason is that the fusion aims to improve segmentation performance. If the framework's performance is worse than that without fusion, it means the fusion is less reliable for facilitating segmentation. Specifically, we utilize energy scores to disseminate reliable fusion. To assess the reliability of depth guidance for segmentation, we first calculate the free energy of each pixel as the energy score for segmentation without EB2F by: $E_{seg}(\mathbf{x})=-\log \sum_{i=1}^K e^{P^{i}(\mathbf{x})}$. Similarly, the free energy of each pixel for segmentation with EB2F is formulated as $\tilde{E}_{seg}(\mathbf{x})=-\log \sum_{i=1}^K e^{\tilde{P}^{i}(\mathbf{x})}$, where $\tilde{P}$ and $P$ represent the predictions with and without EB2F, respectively.

\begin{table*}[t]
  \centering
   \captionsetup{font=small}
   \small
  \resizebox{0.9\linewidth}{!}{ 
  \begin{tabular}{l|cccccccccccccccc|c}
\toprule
Method        & \rotatebox{90}{road}& \rotatebox{90}{s.walk} & \rotatebox{90}{build.} & \rotatebox{90}{wall} & \rotatebox{90}{fence} & \rotatebox{90}{pole} & \rotatebox{90}{light} & \rotatebox{90}{sign} & \rotatebox{90}{veget} & \rotatebox{90}{sky}  & \rotatebox{90}{person} & \rotatebox{90}{rider} & \rotatebox{90}{car}  & \rotatebox{90}{bus}  & \rotatebox{90}{moto.} & \rotatebox{90}{bike} & mIoU \\ \hline\hline
Source \cite{tranheden2021dacs}        & 36.3 & 14.6   & 68.8   & 9.2  & 0.2   & 24.4 & 5.6   & 9.1  & 69.0  & 79.4 & 52.5   & 11.3  & 49.8 & 9.5  & 11.0  & 20.7  & 29.5 \\
R-MRNet \cite{zheng2021rectifying}             & 87.6 & 41.9   & 83.1   & 14.7 & 1.7   & 36.2 & 31.3  & 19.9 & 81.6  & 80.6 & 63.0   & 21.8  & 86.2 & \textbf{40.7} & 23.6  & 53.1  & 47.9 \\
IAST \cite{mei2020instance}                & 81.9 & 41.5   & 83.3   & 17.7 & 4.6   & 32.3 & 30.9  & 28.8 & 83.4  & 85.0 & 65.5   & 30.8  & \textbf{86.5} & 38.2 & 33.1  & 52.7 & 49.8 \\
DACS  \cite{tranheden2021dacs}              & 80.6 & 25.1   & 81.9   & \textbf{21.5} & 2.9   & 37.2 & 22.7  & 24.0 & 83.7  & \textbf{90.8 }& 67.6   & 38.3  & 82.9 & 38.9 & 28.5  & 47.6 & 48.3 \\
GIO-Ada  \cite{ChenLCG19}       & 78.3 & 29.2   & 76.9   & 11.4 & 0.3   & 26.5 & 10.8  & 17.2 & 81.7  & 81.9 & 45.8   & 15.4  & 68.0 & 15.9 & 7.5   & 30.4  & 37.3 \\
DADA  \cite{VuJBCP19}      & 89.2 & 44.8   & 81.4   & 6.8  & 0.3   & 26.2 & 8.6   & 11.1 &\textbf{ 84.8 } & 84.0 & 54.7   & 19.3  & 79.7 & 40.7 & 14.0  & 38.8 &  42.6 \\
CTRL   \cite{SahaOPKCGG21}         & 86.4 & 42.5   & 80.4   & 20.0 & 1.0   & 27.7 & 10.5  & 13.3 & 80.6  & 82.6 & 61.0   & 23.7  & 81.8 & 42.9 & 21.0  & 44.7 & 45.0 \\ 
\hline
CorDA \cite{0013DHGF21}  &87.8&46.7&85.4&20.6&3.1&38.0&38.8&44.0&84.3&85.8&67.9&38.9&84.6&37.5&\textbf{33.7}&54.2&53.2\\
SMART+${\textit{Fusion}_{1}}$&94.2&65.1&85.3&15.3&3.8&39.5&40.3&44.8&83.5&89.0&67.7&\textbf{41.2}&86.2&42.7&20.4&51.7&54.4\\
% 94.0&63.5&85.6&17.8&5.1
% &40.2&39.4&46.8&83.6&88.1&67.9&41.1&85.3&36.6&20.1&53.6&
% 54.3 \\
SMART+${\textit{Fusion}_{2}}$&\textbf{94.6}&\textbf{65.9}&\textbf{86.3}&13.7&\textbf{4.7}&\textbf{40.6}&\textbf{41.8}&\textbf{47.6}&84.3&88.7&\textbf{69.4}&\textbf{41.2}&86.4&35.0&29.9&\textbf{55.2}&\textbf{55.3}\\
\bottomrule
\end{tabular}}
\vspace{-5pt}
  \caption{Semantic segmentation results on the SYNTHIA-to-Cityscapes.}
  \vspace{-10pt}
  \label{tab:synthia}
\end{table*}

To measure the reliability of fusion, we compare the energy scores of the predictions for the same pixel. The larger the free energy, the less reliable the fusion or guidance is. As shown in Fig. \ref{fig:RFA}, the energy score $E_{dep}(\mathbf{x})$ of the prediction $P_{1}$ is larger than $\tilde{P}_{1}$ in the same pixel of an image, which means that the prediction $P_{1}$ is less accurate than $\tilde{P}_{1}$. Therefore, inspired by the online distillation \cite{wang2021knowledge, ZhangXHL18}, we enable the prediction ${P}_{1}$ to be close to $\tilde{P}_{1}$ using KL divergence. Finally, the predictions $P_1$ and $\tilde{P}_2$ of the whole image are reliable, which means that the fusion is more reliable based on collaborative learning. Specifically, we utilize the matrix $m_{(h,w)} \in R^{H \times W}$ to indicate which prediction is more reliable. If the energy score of $\tilde{P}_{seg}$ is less than that of $P_{seg}$, $m_{(h,w)}$ is set to 1, and vice versa for $m_{(h,w)}=0$. Based on the matrix $m_{(h,w)}$, we use the KL divergence to enable the decoders with and without fusion to learn from each other.
{\setlength\abovedisplayskip{2pt}
\setlength\belowdisplayskip{2pt}
\begin{equation}
\scriptsize
\begin{aligned}
\mathcal{L}_{\text{RFA}}^{seg} &= \frac{1}{H \!\times\! W\!-\!M} \sum_{i=1}^{H}\sum_{i=j}^{W} (1\!-\!m_{(h,w)}) \text{KL} \left({{P}_{seg}(\mathbf{x})}_{(h, w)}\|{{\tilde{P}_{seg}}(\mathbf{x})}_{(h, w)}\right) \\
&\!+\! \frac{1}{M} \sum_{i=1}^{H}\sum_{i=j}^{W} m_{(h,w)} \text{KL} \left({{\tilde{P}_{seg}}(\mathbf{x})}_{(h, w)}\|{{P}_{seg}(\mathbf{x})}_{(h, w)}\right), 
\end{aligned}
\end{equation}}
where $M = \sum_{h=1}^{H} \sum_{w=1}^{W} m_{(h,w)}$. 
\begin{figure}
    \centering
     \captionsetup{font=small}
    \includegraphics[width=0.7\linewidth]{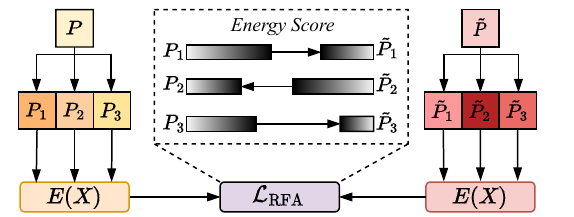}
    %/vspace{-10pt}
    \caption{Illustration of the proposed RFA. $P_1$ and $\tilde{P}_1$ are predictions of same pixel without and with fusion module.}
    \label{fig:RFA}
    \vspace{-20pt}
\end{figure} 
Moreover, we utilize the energy of depth to disseminate reliable fusion with semantics for depth estimation. We utilize the matrix $\hat{m}$ to denote the reliability of depth predictions from our framework with or without fusion. Same as segmentation, if the energy score of $\tilde{P}_{dep}$ is less than that of $P_{dep}$, $m_{(h,w)}$ is set to 1, and vice versa for $m_{(h,w)}=0$.
{\setlength\abovedisplayskip{2pt}
\setlength\belowdisplayskip{2pt}
\begin{equation}
\tiny
\begin{aligned}
\mathcal{L}_{\text{RFA}}^{dep} &= \frac{1}{H \!\times\! W \!-\!\hat{M}} \sum_{i=1}^{H}\sum_{i=j}^{W} (1\!-\!\hat{m}_{(h,w)}) \text{BerHu}\left({P_{dep}(\mathbf{x})}_{(h,w)}\!-\!{\tilde{P}_{dep}(\mathbf{x})}_{(h,w)}\right) \\
&\!+\! \frac{1}{\hat{M}} \sum_{i=1}^{H}\sum_{i=j}^{W} \hat{m}_{(h,w)}\text{BerHu}\left({\tilde{P}_{dep}(\mathbf{x})}_{(h,w)}\!-\!{P_{dep}(\mathbf{x})}_{(h,w)}\right), 
\end{aligned}
\end{equation}}
where $\hat{M} = \sum_{h=1}^{H} \sum_{w=1}^{W} \hat{m}_{(h,w)}$.

In summary, the loss function of RFA is defined as
{\setlength\abovedisplayskip{2pt}
\setlength\belowdisplayskip{2pt}
\begin{equation}
\small
\begin{aligned}
\mathcal{L}_{\text{RFA}}= \mathcal{L}_{\text{RFA}}^{seg}+\alpha\mathcal{L}_{\text{RFA}}^{dep}.\nonumber
\end{aligned}
\end{equation}}
Note that RFA is performed on both source and target domains. The overall objective of the proposed SMART is
{\setlength\abovedisplayskip{2pt}
\setlength\belowdisplayskip{2pt}
\begin{equation}
\small
\begin{aligned}
\mathcal{L}=\mathcal{L}_{s}+\beta\mathcal{L}_{\text{RFA}}, \nonumber
\end{aligned}
\end{equation}}
where $\beta$ is a hyperparameter set to 1.0.

\section{Experiments}

\subsection{Datasets and Implementations.}
We utilize two benchmark tasks, SYNTHIA $\to$ Cityscapes and GTA5 $\to$ Cityscapes, to evaluate the performance of our model. For fair comparison, we choose ResNet-101 as the shared encoder, DeepLabv2 as the task decoder, and residual blocks as the task net, following the approach~\cite{tranheden2021dacs}. 

% \textit{More details about datasets and implementations can be found in suppl. material.}

\subsection{Comparison.}
In the result tables, the source model is trained using the source data and then tested with the target data. Note that `baseline' refers to the entire framework without EB2F and RFA. Additionally, we get the results from \cite{0013DHGF21}. 

\noindent\textbf{Results on GTA5 $\to$ Cityscapes.} 
% We first evaluate the effectiveness of the proposed SMART for learning semantic segmentation from the GTA5 dataset to 
The results are shown in Tab. \ref{tab:gta5}. 
Overall, our SMART achieves mIoU of \textbf{57.7\% }and \textbf{57.0\%} with ${\textit{Fusion}_{1}}$ and ${\textit{Fusion}_{2}}$, respectively. The performance of SMART with two fusion schemes are both better than the SoTA performance with absolute improvements of \textbf{+1.0\%} and \textbf{+0.3\%}. The improvement over previous methods mainly comes from the category of `\textit{fence}'(\textbf{+2.9\%}), `\textit{terrain}'(\textbf{+2.8\%}), `\textit{truck}'(\textbf{+13.5\%}), and `\textit{moto}.' (\textbf{+3.3\%}). The results demonstrate the effectiveness of alleviating the discrepancy between features and ensuring reliable feature fusion. Moreover, we also provide a few qualitative examples in Fig. \ref{fig:visulization}, and can observe that the segmentation quality is largely improved on easily confusing classes.

\begin{figure}[t]
\captionsetup{font=small}
    \centering
    \captionsetup{font=small}
    \includegraphics[width=0.85\linewidth]{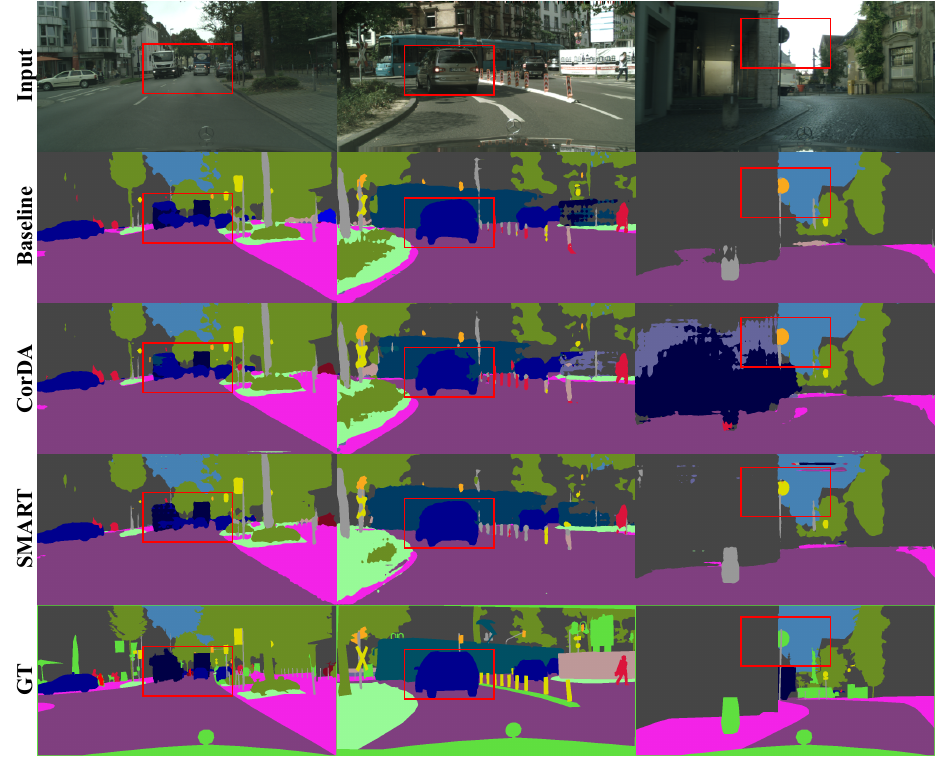}
    \vspace{-5pt}
    \caption{Qualitative results for GTA5-to-Cityscapes.}
    \vspace{-15pt}
    \label{fig:visulization}
\end{figure}

\begin{figure}[t]
    \centering
     \captionsetup{font=small}
    \includegraphics[width=0.87\linewidth]{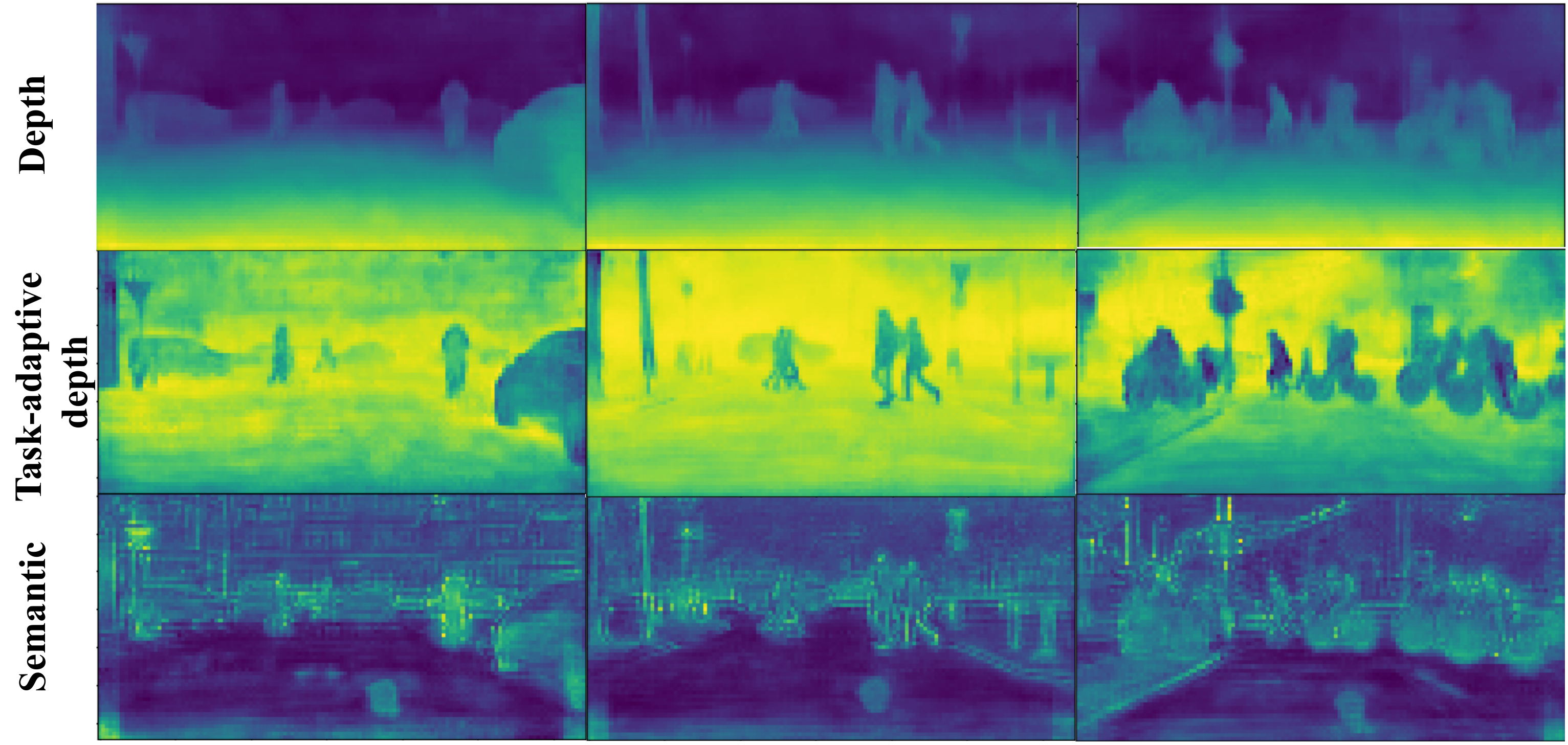}
    \vspace{-5pt}
    \caption{Visualization of features with and without EB2F.}
    \vspace{-20pt}
    \label{fig:EB2F}
\end{figure}
\begin{figure}[t]
    \centering
     \captionsetup{font=small}
    \includegraphics[width=0.9\linewidth]{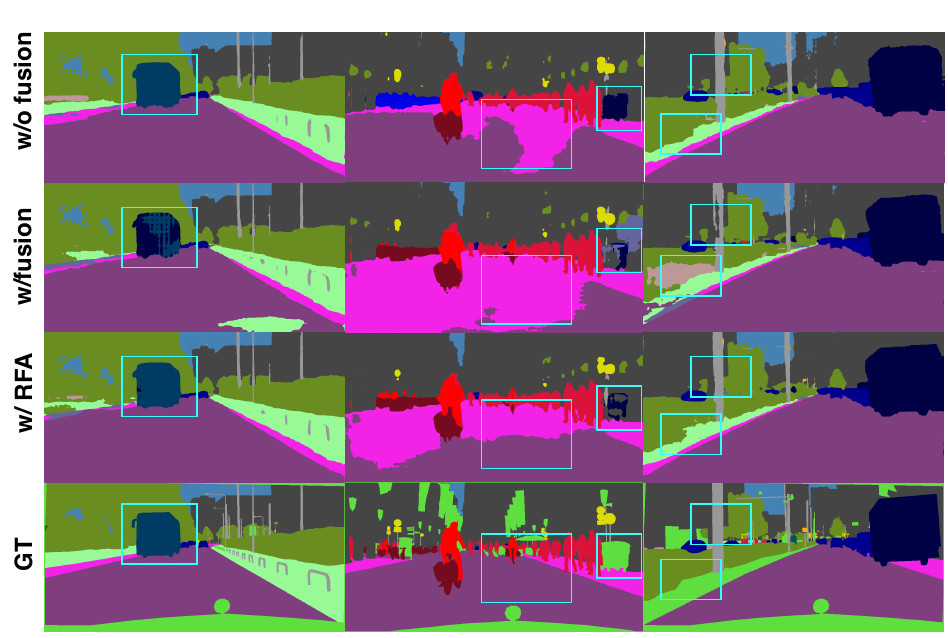}
    \vspace{-5pt}
    \caption{Segmentation results with and without RFA.}
    \label{fig:RFA_results}
    \vspace{-20pt}
\end{figure}

% \noindent \textbf{Feature Visualization.}
% To demonstrate the effectiveness of our proposed SMART, we visualize the feature representation of data before and after utilizing SMART on the Cityscapes dataset by using the t-SNE embedding \cite{DonahueJVHZTD14}. As shown in Fig. \ref{}, 
% {baseline,  hard fusion, CorDA, SMART}

\noindent\textbf{Results on SYNTHIA $\to$ Cityscapes.} To further demonstrate the effectiveness of the proposed SMART, 
% and the importance of explicitly learning task-adaptive features and achieving reliable fusion,
we compare our method with previous works on the SYNTHIA-to-Cityscapes task. Tab. \ref{tab:synthia} shows the experimental results with the common 16 classes. Our method with two schemes yields an absolute improvement of \textbf{1.2\%} and \textbf{2.1\%} mIoU over CorDA, and achieves \textbf{54.4\%}  and \textbf{55.3\%} mIoU, respectively. This outperforms competing methods by a significant margin. Specifically, our SMART with ${\textit{Fusion}_{2}}$ achieves the highest mIoU in \textbf{10} categories. The improvement over previous methods mainly comes from the category of `\textit{road}' (\textbf{+6.8\%}),`\textit{s.walk} (\textbf{+19.2\%})', `\textit{pole}'(\textbf{+2.6\%}), and `\textit{sign}'(\textbf{+3.6\%}). This again verifies the effectiveness of our SMART, which pursues better utilization of depth guidance.

\subsection{Ablation Study and Analysis}
\noindent\textbf{Individual components of SMART.} The main contribution of SMART is to utilize energy estimation to learn task-adaptive features and enable reliable fusion. To validate it, we conduct an ablation study on each of these modules with ${\textit{Fusion}_{1}}$. As shown in Tab. \ref{tab:abl_RFA}, EB2F and RFA improve the baseline results by \textbf{+3.7\%} and \textbf{+1.5\%}, respectively, yielding \textbf{57.1\%} and \textbf{54.9\%} mIoU. The results indicate that EB2F can obtain task-adaptive features for better fusion, while RFA can ensure that the fusion is reliable. Moreover, the integration of these two modules leads to a \textbf{4.3\%} performance gain, resulting in a \textbf{57.7\%} mIoU on the target domain. We also provide a few qualitative examples in Fig. \ref{fig:EB2F} and Fig. \ref{fig:RFA_results} to demonstrate the effectiveness of these two modules. Fig. \ref{fig:EB2F} shows the discrepancy between the depth features with semantic features is effectively reduced after learning task-adaptive depth features via Hopfield energy. Additionally, we can observe in Fig. \ref{fig:RFA_results} that the predictions of model with fusion are more accurate in some regions, \eg, `\textit{bus}' and `\textit{road}', after using RFA module to enable predictions with (w/) fusion to learn from predictions without (w/o) fusion.

\noindent\textbf{Individual modules of EB2F.} To investigate the efficacy of each component of EB2F, we perform ablation studies on GTA5$\rightarrow$ Cityscapes task. The model with $\text{EB2F}_{seg}$ extracts task-adaptive depth features for segmentation, and the model with $\text{EB2F}_{dep}$ extracts task-adaptive semantic features for depth estimation. In Tab. \ref{tab:abl_RFA}, satisfactory and consistent gains from the baseline to our full method demonstrate the effectiveness of each component. Compared with the baseline, the model with task-adaptive semantic and task-adaptive depth features achieves \textbf{56.1\% }and \textbf{54.8\%} mIoU, separately. And combining both task-adaptive features further improves the baseline performance to \textbf{57.1\%} mIoU, demonstrating the effectiveness of extracting task-adaptive features. 

\noindent\textbf{Individual components of RFA.} We evaluate the effectiveness of the components of RFA and investigate how the combination of the two parts contributes to the segmentation performance on the GTA5 $\rightarrow$ Cityscapes task. $\text{RFA}_{seg}$ measures the reliability of feature fusion on the segmentation task, and $\text{RFA}_{dep}$ measures the reliability of feature fusion on the depth estimation task. As shown in Tab. \ref{tab:abl_RFA}, we observe the followings: (1) $\text{RFA}_{seg}$ and $\text{RFA}_{dep}$ led to a \textbf{+1.4\%} and \textbf{+1.0\%} mIoU improvement, respectively, which verifies the effectiveness of assessing the reliability of feature fusion and enabling reliable feature fusion. (2) When combining both losses, our SMART framework further improves the baseline performance to \textbf{54.9\% }mIoU, surpassing the baseline model by a clear margin, thus demonstrating the effectiveness of these two RFA components.
% \begin{table*}[t]
%   \centering
%   \resizebox{0.99\linewidth}{!}{ 
%   \begin{tabular}{|l|c|c|ccccccccccccccccccc|c|}
% \hline
% Baseline &$EB2F_{dep}$ & $EB2F_{seg}$& \rotatebox{90}{road} & \rotatebox{90}{.wallk} & \rotatebox{90}{build.} & \rotatebox{90}{wall} & \rotatebox{90}{fence} & \rotatebox{90}{pole} & \rotatebox{90}{light} & \rotatebox{90}{sign} & \rotatebox{90}{veget.} & \rotatebox{90}{terraim} & \rotatebox{90}{sky}  & \rotatebox{90}{person} & \rotatebox{90}{rider} & \rotatebox{90}{car}  & \rotatebox{90}{truck} & \rotatebox{90}{bus}  & \rotatebox{90}{train} & \rotatebox{90}{moto.}& \rotatebox{90}{bike} & mIoU \\ \hline\hline
% \makecell[c]{\cmark}&& & 93.9 & 64.5    & 87.0   & 26.3 & 40.1 & 41.6 & 46.3  & 50.9 & 87.5  & 46.0    & 88.3& 66.8   & 36.2 & 89.5 & 52.7  & 52.5 & 0.00  & 12.4 & 32.6   &53.4\\
% \makecell[c]{\cmark}&\makecell[c]{\cmark}&\\
% \makecell[c]{\cmark}&&\makecell[c]{\cmark}&91.9&55.3&87.0&38.0&33.6&37.1&43.2&48.7&87.1&47.3&89.9&63.6&27.2&89.4&56.5&55.7&0.0&29.0&53.5&54.4\\
% \makecell[c]{\cmark}&\makecell[c]{\cmark}&\makecell[c]&93.7&58.9&88.1&40.8&42.3&41.1&47.8&50.3&87.8&47.1&87.3&66.2&27.7&91.4&65.4&54.0&0.0&44.1&51.8&57.1
% \\
% \hline
% \end{tabular}}
%   \caption{Ablation study about EB2F}
%   \label{tab:abl_EB2F}
% \end{table*}
\begin{table}[t]
\small
\setlength{\tabcolsep}{4.8mm}
  \centering
   \captionsetup{font=small}
    \resizebox{0.7\linewidth}{!}{
   \small
  \begin{tabular}{l|c|c|c}
\toprule
Baseline &EB2F & RFA & mIoU \\ \hline\hline
\makecell[c]{\cmark}&\makecell[c]{\xmark}&\makecell[c]{\xmark} &53.4\\
\makecell[c]{\cmark}&\makecell[c]{\cmark}&\makecell[c]{\xmark}&57.1
\\
\makecell[c]{\cmark}&\makecell[c]{\xmark}&\makecell[c]{\cmark}&54.9
\\
\makecell[c]{\cmark}&\makecell[c]{\cmark}&\makecell[c]{\cmark}&\textbf{57.7}\\
\hline
\hline
Baseline &${\text{EB2F}}_{seg}$ & ${\text{EB2F}}_{dep}$& mIoU \\ \hline
\makecell[c]{\cmark}&\makecell[c]{\xmark}& \makecell[c]{\xmark}&53.4\\
\makecell[c]{\cmark}&\makecell[c]{\cmark}&\makecell[c]{\xmark}&56.1\\
\makecell[c]{\cmark}&\makecell[c]{\xmark}&\makecell[c]{\cmark}&54.8\\
\makecell[c]{\cmark}&\makecell[c]{\cmark}&\makecell[c]{\cmark}&\textbf{57.1}
\\
\hline\hline
Baseline &$\text{RFA}_{seg}$ & $\text{RFA}_{dep}$& mIoU \\ 
\hline
\makecell[c]{\cmark}&\makecell[c]{\xmark}& \makecell[c]{\xmark}&53.4\\
\makecell[c]{\cmark}&\makecell[c]{\cmark}&\makecell[c]{\xmark}&54.8
\\
\makecell[c]{\cmark}&\makecell[c]{\xmark}&\makecell[c]{\cmark}&54.4\\

\makecell[c]{\cmark}&\makecell[c]{\cmark}&\makecell[c]{\cmark}&\textbf{54.9}
\\
\bottomrule
\end{tabular}}
  \caption{ Ablation study on the individual components of SMART, EB2F, and RFA using GTA5-to-Cityscapes.}
  \vspace{-10pt}
  \label{tab:abl_RFA}
\end{table}

\begin{table}[t]
\setlength{\tabcolsep}{3.8mm}
    \centering
     \captionsetup{font=small}
        \resizebox{0.7\linewidth}{!}{
     \small
    \begin{tabular}{c|ccccc}
    \toprule
         $\beta$&0&0.1&0.5&1.0&2.0  \\
         \hline
         mIoU&53.4&54.4&54.5&\textbf{54.8}&54.2\\
\hline\hline
     $\gamma$&0&0.1&0.2&0.5&1.0 \\
         \hline
         mIoU&53.4&50.8&53.9&55.9&\textbf{57.1}\\
      %   \hline
      %    \hline
      % steps&0&1&2&3&4 \\
      %    \hline
      %    mIoU&53.4&57.1&57.3&&\\
         \bottomrule
    \end{tabular}}
    \caption{Influence of $\beta$ and $\gamma$ using using GTA5-to-Cityscapes.}
    \vspace{-15pt}
    \label{tab:beta}
\end{table}

\noindent\textbf{Influence of $\beta$ and $\gamma$}. We investigate the impact of $\beta$, which represents the weight assigned to the supervised and EFA losses. $\beta$ is varied from 0.1 to 2.0, and the results are presented in Tab. \ref{tab:beta}. Our findings highlight the effectiveness of ensuring reliable feature fusion for segmentation. We select a value of 1.0 for $\beta$ as it demonstrates the best trade-off between the supervised and RFA losses. To evaluate the effectiveness of extracting task-adaptive features, we conduct a hyper-parameter analysis about $\gamma$ in the SMART framework without RFA. $\gamma$ is varied from 0 to 1, and we can observe a gradual improvement in performance. This demonstrates the effectiveness of obtaining task-adaptive features for feature fusion. Notably, when $\gamma = 0$, semantic and depth features are directly added. However, the model's performance achieves the same results as the baseline, suggesting that depth guidance is ineffective for segmentation. 

% \noindent\textbf{Comparing with other orthogonal methods.} This work mainly focuses on using depth guidance for improving segmentation performance in the UDA setting, and the most relevant SOTA method is CorDA. We thus take CorDA as our baseline and demonstrate our ideas on top of it. The UDA for segmentation methods \cite{hoyer2022hrda1, hoyer2022daformer, chen2022smoothing, li2022class, zhang2021prototypical} tackle the problem in another way \textit{not using depth guidance}. The ideas of these methods are \textbf{orthogonal} to ours, thus the proposed method is expected to bring up a complementary effect to further improve their accuracies. In Tab. \ref{tab:GTA5_SOTA}, we further conduct experiments based on Transformer-based HRDA \cite{hoyer2022hrda1} which has shown SOTA accuracies on the GTA5-to-Cityscapes task. Compared to the HRDA methods, our proposed method with two schemes achieves a dramatic increase in mIoU by \textbf{+1.0\%} and \textbf{+1.3\%}, respectively. The results show a significant improvement over HRDA, which demonstrate the effectiveness of our proposed method. \textit{Due to the page limit, more experiments and analyses can be found in the suppl. material.}

\section{Conclusion}
% %/vspace{-5pt}
\label{sec:con}
In this work, we presented a novel energy-based model framework, SMART, for the effective fusion of semantic and depth features in UDA for semantic segmentation with self-supervised depth estimation. Our proposed energy-based feature fusion (EB2F) method generates the task-adaptive semantic and depth features for better feature fusion, while the energy-based reliable fusion assessment (RFA) ensures reliable fusion for improving segmentation performance with depth-guidance. The experimental results demonstrate that our proposed method outperforms SoTA methods by a large margin, highlighting the effectiveness of our approach for semantic segmentation with depth guidance. Moreover, our proposed energy-based framework is a plug-and-play method for domain adaptation and multi-task learning, which has the potential for wide applicability in robotics systems. 

% \noindent\textbf{Limitation and Future Work:} In this paper, we mostly use the Hopfield energy function to build up our SMART framework by measuring the discrepancy between semantic and depth features. Future work will explore other energy functions to better learn task-adaptive features. Moreover, our SMART framework can be extended to explore the correlation between different tasks to improve the performance of these tasks, \eg, depth estimation.

%%%%%%%%%%%%%%%%%%%%%%%%%%%%%%%%%%%%%%%%%%%%%%%%%%%%%%%%%%%%%%%%%%%%%%%%%%%%%%%%
% \clearpage
\bibliographystyle{./IEEEtran}
\bibliography{./IEEEabrv,./ref}
%% \bibliographystyle{./IEEEtran} % use IEEEtran.bst style
%% \bibliography{./IEEEabrv,./IEEEexample}
\end{document}

\maketitle

\thispagestyle{empty}
\pagestyle{empty}

\begin{abstract}

Due to the lack of space in the main paper, we provide more details of the proposed method and experimental results in the supplementary material. Sec.~\ref{discussion} delves into a comprehensive discussion of the proposed SMART.  Sec.~\ref{comparison} shows the comparison between softmax score and energy score. Sec. \ref{dataset} and Sec.~\ref{Implementations} show the details of datasets and implementations. Sec.~\ref{results} shows the details of the experimental results. 
Sec.~\ref{ablation} provides an ablation study about the steps of updating the input pattern and the results. Sec.~\ref{Visulization} provides a visual representation of the results obtained on the GTA5-to-Cityscapes and SYNTHIA-to-Cityscapes datasets. Sec. ~\ref{algorithm} adds the Algorithm of the proposed SMART framework. Sec.~\ref{limitation} introduces the limitation and future work of the proposed SMART.

% Due to the lack of space in the main paper, we provide more details of the proposed method and experimental results in the supplementary material. Sec.~\ref{discussion} delves into a comprehensive discussion of the proposed SMART. Sec.~\ref{dataset} and~\ref{Implementations} meticulously elucidate the dataset specifics and implementation details, respectively. Sec.~\ref{results} meticulously outlines the specifics of the experimental results. Sec.~\ref{ablation} conducts an exhaustive ablation study concerning the steps involved in updating the input pattern and presents corresponding outcomes. Sec.~\ref{Visulization} offers a visual representation of the results obtained on the GTA5-to-Cityscapes and SYNTHIA-to-Cityscapes datasets. Section~\ref{comparison} systematically compares the softmax score with the energy score. Section~\ref{limitation} succinctly introduces the limitations and outlines potential avenues for future research within the proposed SMART framework. Section~\ref{algorithm} supplements the manuscript with the algorithm detailing the SMART framework. Finally, Section~\ref{code} provides a comprehensive code implementation of critical modules for those interested in reproducing or building upon our work.

\end{abstract}

%%%%%%%%% BODY TEXT

\section{Discussion}
\label{discussion}
\noindent\textbf{Difference between our proposed method and UDA for semantic segmentation.} 

This work mainly focuses on using depth guidance for improving segmentation performance in the UDA setting, and the most relevant SOTA method is CorDA. We thus take CorDA as our baseline and demonstrate our ideas on top of it. The UDA for segmentation methods \cite{hoyer2022hrda1, hoyer2022daformer, chen2022smoothing} tackle the problem in another way \textit{not using depth guidance}. The ideas of these methods are \textbf{orthogonal} to ours, thus the proposed method is expected to bring up a complementary effect to further improve their accuracy.

\noindent\textbf{Difference between cross-attention and our proposed energy-based fusion.} 

EBMs are adopted as a \textit{unified} framework or an \textit{effective} tool to obtain task-adaptive feature fusion and assess the quality of the feature fusion for enhancing segmentation with depth guidance. Unlike cross-attention modules which only focus on merging features, we first propose to obtain specific \textit{task-adaptive} features (depth) for semantics features,  then use a cross-attention module or direct addition to fuse them. Moreover, we can utilize Hopfield energy to explicitly measure the discrepancy between depth and semantic features, and utilize $\gamma$ and iteration in Eq.8 to control the degree of updating features (depth) for better fusion with other features (semantics). The ablation study about these two parameters is shown in Tabs. \ref{tab:gamma} and \ref{tab:update}.

\noindent\textbf{Details of RFA.}

Previous works have shown that energy score is more effective than softmax which is utilized for calculating entropy for detecting out-of-distribution data. Moreover, this work not just focuses on the selection, and we propose to enable predictions with or without fusion \textit{bidirectionally}, which learns from each other based on collaborative learning, and make both predictions more accurate. Due to that the target data is unlabeled in the training and inference, we propose to utilize unsupervised energy score to disseminate reliable fusion instead of using supervised GT. 

Depth guidance has been shown to improve the segmentation accuracy in UDA settings, thanks to the geometric cues it brings, in literature. Therefore, the predictions with fusion are generally \textit{more accurate} than the predictions without fusion. However, note that we tackle the dense prediction problem, i.e., predicting in each individual pixel. Given a single image instance, \textbf{not all pixels} might have accurate depth estimation, thus we better consider the feature fusion selectively to pixels. That is achieved by the proposed RFA, which enables predictions with or without fusion bidirectionally, letting them learn from each other in pixel-wise level. Therefore, $W_d^1$ after learning is not zero. As shown in Tabs. 3 and 4 in the main paper, this bidirectional module with $W_d^1$ improves the baseline accuracy.

\noindent\textbf{Details of training and inference.}

The inputs for training are labeled source and unlabeled target data, and the inputs for inference are only the unlabeled target data. In the training, we alleviate the domain gap by gradually obtaining reliable pseudo labels of target data and then reducing the cross-entropy loss, which is a self-supervised pseudo-labeling approach and commonly used for domain alignment in previous works. During the training, we measure the discrepancy between predictions \textit{with and without} the feature fusion as part of our loss functions. During the inference, we use \textit{only }the pipeline that goes through the feature fusion.

\noindent\textbf{Illustration of EB2F module.}

\begin{figure}[t]
    \centering
     \captionsetup{font=small}
    \includegraphics[width=0.9\linewidth]{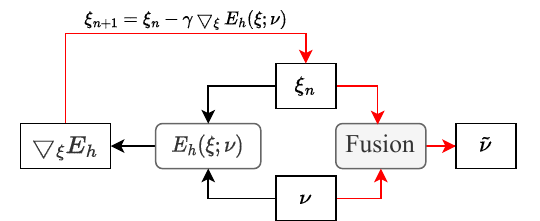}
    \vspace{-5pt}
    \caption{The illustration of the proposed EB2F module.}
    \label{fig:fusion}
    \vspace{-15pt}
\end{figure}

The overview of the proposed EB2F module is shown as Fig. \ref{fig:fusion}. The black arrow means the process of getting the gradient $\nabla_{\boldsymbol{\xi}}$, and the inputs for the fusion are an updated pattern $\xi_{n}$ and another $\mu$. The loop means that we can update the pattern $\xi$ by $n$ iterations to obtain better task-adaptive features.

\noindent\textbf{About improvement of energy-based module.}

Our method is agnostic to fusion models, and we conduct experiments using different fusion schemes (Scheme 1 and Scheme 2). We \textit{directly }add features (\textbf{$Fusion_{1}$}) without any fusion module, and the results about $Fusion_{1}$ and $EB2F$ on Tabs. 1, 2, 3, 4 in the main paper demonstrate the effectiveness of energy design.

\noindent\textbf{Definition of E(x).}

E(x) is the \textbf{general} energy function that maps the variable x to a single non-probabilistic scalar. $E_{seg}$ and $E_{dep}$ are the specific energy functions for segmentation (classification) and depth estimation (regression) tasks, respectively. And $E_{seg}$ and $E_{dep}$ are calculated using the outputs of networks and ground truth.

\noindent\textbf{Difference between uncertainty estimation and EBMs.}

Uncertainty estimation and EBMs are two distinct concepts. Uncertainty estimation quantifies the uncertainty or lack of confidence in predictions, and EBMs assign the lowest energy to correct predictions and higher energy values to other incorrect predictions.

\section{Comparison between Softmax Score and Energy Score}
\label{comparison}

Note that the comparison is provided from~\cite{LiuWOL20}. Energy score can serve as a simple and effective replacement for the softmax confidence score~\cite{hendrycks2016baseline} for any pre-trained neural network. To establish this, we derive a mathematical connection between the energy score and the softmax confidence score:
$$
\begin{aligned}
\max _y p(y \mid \mathbf{x}) & =\max _y \frac{e^{f_y(\mathbf{x})}}{\sum_i e^{f_i(\mathbf{x})}}=\frac{e^{f^{\max }(\mathbf{x})}}{\sum_i e^{f_i(\mathbf{x})}} \\
& =\frac{1}{\sum_i e^{f_i(\mathbf{x})-f^{\max }(\mathbf{x})}} \\
\Longrightarrow \log \max _y p(y \mid \mathbf{x}) & =E\left(\mathbf{x} ; f(\mathbf{x})-f^{\max }(\mathbf{x})\right)\\
&=E(\mathbf{x} ; f)+f^{\max }(\mathbf{x}),
\end{aligned}
$$
This equation shows that the log of the softmax confidence score is equivalent to a special case of the free energy score, where all the logits are shifted by their maximum logit value. When $T=1$, the in-distribution data tends to have a lower value for $E(\mathbf{x} ; f)$ and a higher value for $f^{\max }(\mathbf{x})$. As a result, the shifting leads to a biased scoring function that is no longer proportional to the probability density $p(\mathbf{x})$ for $\mathbf{x} \in \mathbb{R}^D$:
$$
\begin{aligned}
& \log \max _y p(y \mid \mathbf{x})=-\log p(\mathbf{x})+\underbrace{f^{\max }(\mathbf{x})-\log Z}_{\text {Not constant. Larger for in-dist } \mathbf{x}}. \\
\end{aligned}
$$
Unlike the energy score, which is well-aligned with density $p(x)$, the softmax confidence score \textbf{is less reliable} in distinguishing between in- and out-of-distribution examples.
Based on the aforementioned comparison, we conduct a subtle analysis of the energy score of predictions made with and without feature fusion. This is because the feature fusion process is intended to improve the segmentation performance, and if the performance of the framework is worse with feature fusion than without it, then it indicates that the feature fusion is less reliable in facilitating segmentation. To accomplish this, we use energy scores to evaluate the reliability of feature fusion. For a detailed explanation of energy scores, please refer to ~\cite{LiuWOL20}.

\section{Datasets}
\label{dataset}
In this study, we utilize two standard benchmark tasks, SYNTHIA $\to$ Cityscapes and GTA5 $\to$ Cityscapes, to evaluate the performance of our model. We utilize SYNTHIA \cite{ros2016synthia} and GTA5 \cite{richter2016playing} datasets as synthetic source domains, and Cityscapes \cite{cordts2016cityscapes} dataset as the real target domain. To ensure a fair comparison, we generate the depth ground-truths for Cityscapes and GTA5 datasets as CorDA. The \textbf{Cityscapes} dataset is a real-world dataset with densely annotated semantic segmentation for 19 classes, consisting of 2975 images in the training set and 500 images in the validation set, each with a fixed resolution of 2048$\times$1024 pixels. Following the experimental protocol used by \cite{ChenLCG19}, we resize the original images to 1024$\times$512 pixels. The \textbf{GTA5 }dataset is a synthetic dataset generated from a game environment, with 19 classes using Cityscapes-style annotation. It contains 24,966 synthetic images. The \textbf{SYNTHIA }dataset is a synthetic dataset of a virtual road scene, with 16 overlapping classes using Cityscapes-style annotations. We utilize the SYNTHIA-RAND-CITYSCAPES split and 9,400 synthetic images. The source depth supervision is provided by the simulated depth of the dataset. We evaluate the performance of our proposed model using the Intersection Over Union (IoU) for per-class performance, as well as the mean Intersection over Union (mIoU) over all classes.
\section{Implementations}
\label{Implementations}
For the purpose of a fair comparison, we have chosen ResNet-101 as the shared encoder, DeepLabv2 as the task decoder, and residual blocks as the task net, following the approach in \cite{tranheden2021dacs}. We set the batch size to 2 and the learning rate to start at $2.5 \times 10^{-4}$, with a polynomial decay of exponent 0.9. Images from the source domain are scaled to 1280 × 760, while those from the target domain have a resolution of 1024 × 512, which is used as input for training. To further augment the data, we apply random crops of size 512 × 512. We assign weights for the source depth loss and the target depth loss to $\alpha$ = 0.001 as \cite{tranheden2021dacs}. All models undergo training for 250,000 iterations. To ensure more reliable fusion, we further train the models for an additional 50,000 iterations using RFA based on the previously trained model, with the learning rate set to $2.5 \times 10^{-5}$.

\section{Results}
\label{results}
\noindent\textbf{Comparison with previous methods.} 

Tab.~\ref{tab:gta5} shows the detailed comparison results on GTA5-to-Cityscapes.

\begin{table*}[h]
  \centering
  \captionsetup{font=small}
  \resizebox{0.99\linewidth}{!}{ 
  \begin{tabular}{l|ccccccccccccccccccc|c}
\toprule
Method      & \rotatebox{90}{road} & \rotatebox{90}{.walk} & \rotatebox{90}{build.} & \rotatebox{90}{wall} & \rotatebox{90}{fence} & \rotatebox{90}{pole} & \rotatebox{90}{light} & \rotatebox{90}{sign} & \rotatebox{90}{veget.} & \rotatebox{90}{terrain} & \rotatebox{90}{sky}  & \rotatebox{90}{person} & \rotatebox{90}{rider} & \rotatebox{90}{car}  & \rotatebox{90}{truck} & \rotatebox{90}{bus}  & \rotatebox{90}{train} & \rotatebox{90}{moto.}& \rotatebox{90}{bike} & mIoU \\ \hline\hline
Source only\cite{tsai2018learning}  & 75.8 & 16.8    & 77.2   & 12.5 & 21.0  & 25.5 & 30.1  & 20.1 & 81.3   & 24.6    & 70.3 & 53.8   & 26.4  & 49.9 & 17.2  & 25.9 & 6.5   & 25.3  & 36.0   & 36.6 \\
ROAD   \cite{chen2018road}     & 76.3 & 36.1    & 69.6   & 28.6 & 22.4  & 28.6 & 29.3  & 14.8 & 82.3   & 35.3    & 72.9 & 54.4   & 17.8  & 78.9 & 27.7  & 30.3 & 4.0   & 24.9  & 12.6   & 39.4 \\
OutputAdapt \cite{tsai2018learning} & 86.5 & 36.0    & 79.9   & 23.4 & 23.3  & 23.9 & 35.2  & 14.8 & 83.4   & 33.3    & 75.6 & 58.5   & 27.6  & 73.7 & 32.5  & 35.4 & 3.9   & 30.1  & 28.1   & 42.4 \\
ADVENT \cite{vu2019advent}     & 87.6 & 21.4    & 82.0   & 34.8 & 26.2  & 28.5 & 35.6  & 23.0 & 84.5   & 35.1    & 76.2 & 58.6   & 30.7  & 84.8 & 34.2  & 43.4 & 0.4   & 28.4  & 35.3   & 44.8 \\
CBST    \cite{zou2018unsupervised}       & 91.8 & 53.5    & 80.5   & 32.7 & 21.0  & 34.0 & 28.9  & 20.4 & 83.9   & 34.2    & 80.9 & 53.1   & 24.0  & 82.7 & 30.3  & 35.9 & 16.0  & 25.9  & 42.8   & 45.9 \\
BDL   \cite{li2019bidirectional}      & 91.0 & 44.7    & 84.2   & 34.6 & 27.6  & 30.2 & 36.0  & 36.0 & 85.0   & 43.6    & 83.0 & 58.6   & 31.6  & 83.3 & 35.3  & 49.7 & 3.3   & 28.8  & 35.6   & 48.5 \\
MRKLD-SP \cite{zou2019confidence}   & 90.8 & 46.0    & 79.9   & 27.4 & 23.3  & \textbf{42.3} & 46.2  & 40.9 & 83.5   & 19.2    & 59.1 & 63.5   & 30.8  & 83.5 & 36.8  & 52.0 & 28.0  & 36.8  &  46.4  & 49.2 \\
Kim et al. \cite{kim2020learning} & 92.9 & 55.0    & 85.3   & 34.2 & 31.1  & 34.9 &  40.7     &  34.0    &  85.2      &  40.1       &  87.1    &   61.0     &   31.1    &  82.5    &  32.3     &  42.9    &   0.3    &   36.4    & 46.1    &50.2 \\
CAG-UDA \cite{zhang2019category}    & 90.4 & 51.6    & 83.8   & 34.2 & 27.8  & 38.4 &   25.3    &    48.4  &     85.4   &   38.2      &   78.1   &    58.6    &   34.6    &    84.7  &   21.9    &   42.7   &  \textbf{  41.1}   &   29.3    &  37.2   &50.2 \\
FDA  \cite{yang2020fda}        & 92.5 & 53.3    & 82.4   & 26.5 & 27.6  & 36.4 &   40.6    &    38.9  &   82.3     &    39.8     &  78.0    &    62.6    &    34.4   &   84.9   &  34.1     &  53.1    &  16.9     &   27.7    &   46.4  & 50.5\\
PIT   \cite{lv2020cross}      &   87.5   &     43.4    &   78.8     &   31.2   &  30.2     &  36.3    &   39.9    &   42.0   &   79.2     &    37.1     &  79.3    &   65.4     &    \textbf{37.5}   &   83.2   &    46.0   &   45.6   &  25.7     &  23.5     &  49.9   & 50.6\\
IAST   \cite{mei2020instance}      &   93.8   &    57.8     &   85.1     &  \textbf{39.5}    &   26.7    &   26.2   &   43.1    &   34.7   &   84.9     &    32.9     &   88.0   &    62.6    &   29.0    &   87.3   &  39.2     &   49.6   &   23.2    &   34.7    &  39.6  & 51.5  \\ 
DACS    \cite{tranheden2021dacs}      &    89.9  &     39.7    &   87.9     &   30.7   &  39.5   &  38.5    &    46.4   &   52.8   &   \textbf{ 88.0 }   &     44.0    &  88.8    & \textbf{  67.2}     &    35.8   &   84.5   &   45.7    &   50.2   &  0.0     &   27.3   & 34.0 & 52.1\\ 
\hline
CorDA    \cite{0013DHGF21} &\textbf{95.2}&\textbf{65.1}&87.4&33.0&40.0&39.9&47.7&\textbf{53.6}&87.6&45.3&88.7&66.2&33.5&89.9&53.9&53.3&0.2&41.0&\textbf{55.8}&56.7\\
SMART+${\textit{Fusion}_{1}}$& 94.3&59.7&\textbf{88.4}&37.9&\textbf{42.9}&40.0&48.4&51.5&87.3&\textbf{48.1}&88.8&66.6&28.8&\textbf{91.5}&\textbf{67.4}&\textbf{57.9}&0.0&\textbf{44.3}&51.9&\textbf{57.7}\\
SMART+${\textit{Fusion}_{2}}$& 94.5&60.3&88.3&31.8&42.8&41.0&\textbf{48.5}&50.5&87.3&47.3&\textbf{89.0}&66.4&28.0&91.4&63.9&56.1&0.0&44.0&51.2
&57.0
\\
\bottomrule
\end{tabular}}
  \caption{Semantic segmentation results on the GTA5-to-Cityscapes.}
  \label{tab:gta5}
  \vspace{-10pt}
\end{table*}

Tab.~\ref{tab:synthia} shows the detailed comparison results on SYNTHIA-to-Cityscapes.

\begin{table*}[t]
  \centering
   \captionsetup{font=small}
   \small
  \resizebox{0.99\linewidth}{!}{ 
  \begin{tabular}{l|cccccccccccccccc|c}
\toprule
Method        & \rotatebox{90}{road}& \rotatebox{90}{s.walk} & \rotatebox{90}{build.} & \rotatebox{90}{wall} & \rotatebox{90}{fence} & \rotatebox{90}{pole} & \rotatebox{90}{light} & \rotatebox{90}{sign} & \rotatebox{90}{veget} & \rotatebox{90}{sky}  & \rotatebox{90}{person} & \rotatebox{90}{rider} & \rotatebox{90}{car}  & \rotatebox{90}{bus}  & \rotatebox{90}{moto.} & \rotatebox{90}{bike} & mIoU \\ \hline\hline
Source \cite{tranheden2021dacs}        & 36.3 & 14.6   & 68.8   & 9.2  & 0.2   & 24.4 & 5.6   & 9.1  & 69.0  & 79.4 & 52.5   & 11.3  & 49.8 & 9.5  & 11.0  & 20.7  & 29.5 \\
ADVENT \cite{vu2019advent}              & 85.6 & 42.2   & 79.7   & 8.7  & 0.4   & 25.9 & 5.4   & 8.1  & 80.4  & 84.1 & 57.9   & 23.8  & 73.3 & 36.4 & 14.2  & 33.0  & 41.2 \\
CBST   \cite{zou2018unsupervised}              & 68.0 & 29.9   & 76.3   & 10.8 & 1.4   & 33.9 & 22.8  & 29.5 & 77.6  & 78.3 & 60.6   & 28.3  & 81.6 & 23.5 & 18.8  & 39.8  & 42.6 \\
R-MRNet \cite{zheng2021rectifying}             & 87.6 & 41.9   & 83.1   & 14.7 & 1.7   & 36.2 & 31.3  & 19.9 & 81.6  & 80.6 & 63.0   & 21.8  & 86.2 & \textbf{40.7} & 23.6  & 53.1  & 47.9 \\
IAST \cite{mei2020instance}                & 81.9 & 41.5   & 83.3   & 17.7 & 4.6   & 32.3 & 30.9  & 28.8 & 83.4  & 85.0 & 65.5   & 30.8  & \textbf{86.5} & 38.2 & 33.1  & 52.7 & 49.8 \\
DACS  \cite{tranheden2021dacs}              & 80.6 & 25.1   & 81.9   & \textbf{21.5} & 2.9   & 37.2 & 22.7  & 24.0 & 83.7  & \textbf{90.8 }& 67.6   & 38.3  & 82.9 & 38.9 & 28.5  & 47.6 & 48.3 \\
SPIGAN  \cite{lee2018spigan}        & 71.1 & 29.8   & 71.4   & 3.7  & 0.3   & 33.2 & 6.4   & 15.6 & 81.2  & 78.9 & 52.7   & 13.1  & 75.9 & 25.5 & 10.0  & 20.5 & 36.8 \\
GIO-Ada  \cite{ChenLCG19}       & 78.3 & 29.2   & 76.9   & 11.4 & 0.3   & 26.5 & 10.8  & 17.2 & 81.7  & 81.9 & 45.8   & 15.4  & 68.0 & 15.9 & 7.5   & 30.4  & 37.3 \\
DADA  \cite{VuJBCP19}      & 89.2 & 44.8   & 81.4   & 6.8  & 0.3   & 26.2 & 8.6   & 11.1 &\textbf{ 84.8 } & 84.0 & 54.7   & 19.3  & 79.7 & 40.7 & 14.0  & 38.8 &  42.6 \\
CTRL   \cite{SahaOPKCGG21}         & 86.4 & 42.5   & 80.4   & 20.0 & 1.0   & 27.7 & 10.5  & 13.3 & 80.6  & 82.6 & 61.0   & 23.7  & 81.8 & 42.9 & 21.0  & 44.7 & 45.0 \\ 
% CorDA (mono) \cite{0013DHGF21}  &   \checkmark    & 90.2 & 47.5   & 85.6   & 24.5 & 3.0   & 38.2 & 41.6  & 36.5 & 85.9  & 91.7 & 70.3   & 42.4  & 86.0 & 42.9 & 34.7  & 50.4 & 62.0 & 54.5 \\
\hline
CorDA \cite{0013DHGF21}  &87.8&46.7&85.4&20.6&3.1&38.0&38.8&44.0&84.3&85.8&67.9&38.9&84.6&37.5&\textbf{33.7}&54.2&53.2\\
SMART+${\textit{Fusion}_{1}}$&94.2&65.1&85.3&15.3&3.8&39.5&40.3&44.8&83.5&89.0&67.7&\textbf{41.2}&86.2&42.7&20.4&51.7&54.4\\
% 94.0&63.5&85.6&17.8&5.1
% &40.2&39.4&46.8&83.6&88.1&67.9&41.1&85.3&36.6&20.1&53.6&
% 54.3 \\
SMART+${\textit{Fusion}_{2}}$&\textbf{94.6}&\textbf{65.9}&\textbf{86.3}&13.7&\textbf{4.7}&\textbf{40.6}&\textbf{41.8}&\textbf{47.6}&84.3&88.7&\textbf{69.4}&\textbf{41.2}&86.4&35.0&29.9&\textbf{55.2}&\textbf{55.3}\\
\bottomrule
\end{tabular}}
  \caption{Semantic segmentation results on the SYNTHIA-to-Cityscapes.}
  \label{tab:synthia}
\end{table*}

\noindent\textbf{Effectiveness of two components of EBG.} 

Tab.~\ref{tab:abl_EB2F} shows the detailed results of ablation study of the proposed EB2F.

\begin{table*}[t]
  \centering
  \resizebox{0.99\linewidth}{!}{ 
  \begin{tabular}{|l|c|c|ccccccccccccccccccc|c|}
\hline
Baseline &$EB2F_{dep}$ & $EB2F_{seg}$& \rotatebox{90}{road} & \rotatebox{90}{.wallk} & \rotatebox{90}{build.} & \rotatebox{90}{wall} & \rotatebox{90}{fence} & \rotatebox{90}{pole} & \rotatebox{90}{light} & \rotatebox{90}{sign} & \rotatebox{90}{veget.} & \rotatebox{90}{terraim} & \rotatebox{90}{sky}  & \rotatebox{90}{person} & \rotatebox{90}{rider} & \rotatebox{90}{car}  & \rotatebox{90}{truck} & \rotatebox{90}{bus}  & \rotatebox{90}{train} & \rotatebox{90}{moto.}& \rotatebox{90}{bike} & mIoU \\ \hline\hline
\makecell[c]{\cmark}&& & 93.9 & 64.5    & 87.0   & 26.3 & 40.1 & 41.6 & 46.3  & 50.9 & 87.5  & 46.0    & 88.3& 66.8   & 36.2 & 89.5 & 52.7  & 52.5 & 0.0  & 12.4 & 32.6   &53.4\\
\makecell[c]{\cmark}&\makecell[c]{\cmark}&&93.5&57.3&88.4&39.8&44.8&40.6&48.3&54.3&88.2&50.6&90.4&66.0&34.7&90.9&60.2&55.6&0.0&18.5&43.3&56.1
\\
\makecell[c]{\cmark}&&\makecell[c]{\cmark}&93.3&54.8&88.3&31.7&44.1&39.2&48.5&57.5&87.5&50.2&89.0&65.7&33.7&89.4&56.4&47.4&0.0&18.3&46.4&54.8
\\
\makecell[c]{\cmark}&\makecell[c]{\cmark}&\makecell[c]&93.7&58.9&88.1&40.8&42.3&41.1&47.8&50.3&87.8&47.1&87.3&66.2&27.7&91.4&65.4&54.0&0.0&44.1&51.8&57.1
\\
\hline
\end{tabular}}
  \caption{Ablation study about EB2F}
  \label{tab:abl_EB2F}
\end{table*}

Tab.~\ref{tab:abl_RFA} shows the detailed results of ablation study of the proposed RFA.

\begin{table*}[t]
  \centering
  \resizebox{0.99\linewidth}{!}{ 
  \begin{tabular}{|l|c|c|ccccccccccccccccccc|c|}
\hline
Baseline &$RFA_{seg}$ & $RFA_{dep}$& \rotatebox{90}{road} & \rotatebox{90}{.wallk} & \rotatebox{90}{build.} & \rotatebox{90}{wall} & \rotatebox{90}{fence} & \rotatebox{90}{pole} & \rotatebox{90}{light} & \rotatebox{90}{sign} & \rotatebox{90}{veget.} & \rotatebox{90}{terraim} & \rotatebox{90}{sky}  & \rotatebox{90}{person} & \rotatebox{90}{rider} & \rotatebox{90}{car}  & \rotatebox{90}{truck} & \rotatebox{90}{bus}  & \rotatebox{90}{train} & \rotatebox{90}{moto.}& \rotatebox{90}{bike} & mIoU \\ \hline\hline
\makecell[c]{\cmark}&& &93.9 & 64.5    & 87.0   & 26.3 & 40.1 & 41.6 & 46.3  & 50.9 & 87.5  & 46.0    & 88.3& 66.8   & 36.2 & 89.5 & 52.7  & 52.5 & 0.0  & 12.4 & 32.6   &53.4\\
\makecell[c]{\cmark}&\makecell[c]{\cmark}&&94.6&64.1&88.1&37.6&42.8&41.4& 47.7&51.4&87.8&47.5&87.8&67.5&37.2&89.9&52.4&57.0& 0.0&12.0&34.1&54.8
\\
\makecell[c]{\cmark}&&\makecell[c]{\cmark}&94.8&65.0&88.0&35.4&40.5&41.0&47.4&51.9&87.9&44.9&87.4&66.7&36.2&90.0&50.0&58.5&0.0&12.6&35.8&54.4\\

\makecell[c]{\cmark}&\makecell[c]{\cmark}&\makecell[c]{\cmark}&94.7&61.9&86.8&31.7&34.6&39.3&45.2&51.8&86.9&43.5&88.5&63.8&32.3&90.2&53.0&47.6&0.0&38.4&55.2&54.9
\\
\hline
\end{tabular}}
  \caption{ Ablation study about RFA.}
  \label{tab:abl_RFA}
\end{table*}

Tab.~\ref{tab:beta} shows the detailed results about influence of $\beta$.

\begin{table*}[t]
  \centering
  \resizebox{0.99\linewidth}{!}{ 
  \begin{tabular}{|l|ccccccccccccccccccc|c|}
\hline
 & \rotatebox{90}{road} & \rotatebox{90}{.wallk} & \rotatebox{90}{build.} & \rotatebox{90}{wall} & \rotatebox{90}{fence} & \rotatebox{90}{pole} & \rotatebox{90}{light} & \rotatebox{90}{sign} & \rotatebox{90}{veget.} & \rotatebox{90}{terraim} & \rotatebox{90}{sky}  & \rotatebox{90}{person} & \rotatebox{90}{rider} & \rotatebox{90}{car}  & \rotatebox{90}{truck} & \rotatebox{90}{bus}  & \rotatebox{90}{train} & \rotatebox{90}{moto.}& \rotatebox{90}{bike} & mIoU \\ \hline\hline
0& 93.9 & 64.5    & 87.0  & 26.3& 40.1 & 41.6 & 46.3  & 50.9 & 87.5 & 46.0    & 88.3& 66.8   & 36.2 & 89.5 & 52.7  & 52.5 & 0.0  & 12.4& 32.6   &53.4\\
0.1&95.1&64.4&88.1&28.5&42.1&42.5&48.2&50.4&88.0&48.1&88.5&67.3&36.2&90.2&54.9&58.1&0.0&12.3&33.2&54.4
\\
0.5&95.0&64.7&87.6&27.0&39.9&40.5&47.2&50.8&87.9&47.4&88.8&67.3&37.1&89.9&54.2&58.9&0.0&14.4&37.4&54.5
\\
1.0&94.6&64.1&88.1&37.6&42.8&41.4& 47.7&51.4&87.9&47.5&87.8&67.5&37.2&89.9&52.4&57.0& 0.0&12.0&34.1&54.8\\
2.0&95.4&67.5&87.6&26.5&39.0&39.1&47.2&50.8&88.0&47.7&88.7&67.2&37.5&90.2&54.0&58.1&0.0&12.2&33.1&54.2\\
\hline
\end{tabular}}
  \caption{Influence of $\beta$.}
  \label{tab:beta}
\end{table*}
\begin{table*}[t]
  \centering
  \resizebox{0.99\linewidth}{!}{ 
  \begin{tabular}{|l|ccccccccccccccccccc|c|}
\hline
 & \rotatebox{90}{road} & \rotatebox{90}{.wallk} & \rotatebox{90}{build.} & \rotatebox{90}{wall} & \rotatebox{90}{fence} & \rotatebox{90}{pole} & \rotatebox{90}{light} & \rotatebox{90}{sign} & \rotatebox{90}{veget.} & \rotatebox{90}{terraim} & \rotatebox{90}{sky}  & \rotatebox{90}{person} & \rotatebox{90}{rider} & \rotatebox{90}{car}  & \rotatebox{90}{truck} & \rotatebox{90}{bus}  & \rotatebox{90}{train} & \rotatebox{90}{moto.}& \rotatebox{90}{bike} & mIoU \\ \hline\hline
0& 93.9 & 64.5    & 87.0   & 26.3 & 40.1 & 41.6 & 46.3  & 50.9 & 87.5  & 46.0    & 88.3& 66.8   & 36.2 & 89.5 & 52.7  & 52.5 & 0.0  & 12.4 & 32.6   &53.4\\
0.1&92.8&48.9&87.7&32.0&36.0&37.4&47.1&41.9&87.0&46.9&89.6&65.8&36.8&90.5&43.5&45.1&0.0&10.8&25.3&50.8
 \\
0.2&95.0&64.7&88.1&31.8&38.0&37.6&48.2&54.9&88.1&49.7&90.3&65.8&35.8&90.3&53.5&48.3&0.0&12.7&31.3&53.9
\\
0.5&93.4&54.9&88.3&39.4&43.9&41.5&48.1&51.9&88.3&50.7&89.7&66.7&36.7&91.0&60.0&55.7&0.0&18.6&44.8&55.9
\\
1.0&93.7&58.9&88.1&40.8&42.3&41.1&47.8&50.3&87.8&47.1&87.3&66.2&27.7&91.4&65.4&54.0&0.0&44.1&51.8&57.1\\
\hline
\end{tabular}}
  \caption{Inluence of $\gamma$.}
  \label{tab:gamma}
\end{table*}
\begin{table*}
  \centering
  \resizebox{0.99\linewidth}{!}{ 
  \begin{tabular}{|l|ccccccccccccccccccc|c|}
\hline
 & \rotatebox{90}{road} & \rotatebox{90}{.wallk} & \rotatebox{90}{build.} & \rotatebox{90}{wall} & \rotatebox{90}{fence} & \rotatebox{90}{pole} & \rotatebox{90}{light} & \rotatebox{90}{sign} & \rotatebox{90}{veget.} & \rotatebox{90}{terraim} & \rotatebox{90}{sky}  & \rotatebox{90}{person} & \rotatebox{90}{rider} & \rotatebox{90}{car}  & \rotatebox{90}{truck} & \rotatebox{90}{bus}  & \rotatebox{90}{train} & \rotatebox{90}{moto.}& \rotatebox{90}{bike} & mIoU \\ \hline\hline
0& 93.9 & 64.5    & 87.0   & 26.3 & 40.1 & 41.6 & 46.3  & 50.9 & 87.5  & 46.0    & 88.3& 66.8   & 36.2 & 89.5 & 52.7  & 52.5 & 0.0  & 12.4 & 32.6   &53.4\\
1&93.7&58.9&88.1&40.8&42.3&41.1&47.8&50.3&87.8&47.1&87.3&66.2&27.7&91.4&65.4&54.0&0.0&44.1&51.8&57.1\\
2&93.9&57.3&88.1&35.8&41.2&40.3&48.4&51.2&87.7&46.0&89.4&67.4&36.0&91.3&58.1&55.2&0.0&45.7&55.9&57.3
\\
3&95.0&65.1&88.5&39.5&42.4&41.1&48.4&51.8&88.3&48.5&90.7&67.4&36.6&91.2&51.5&46.4&0.0&11.0&23.2&54.0
\\
4&88.4&42.2&82.7&26.5&26.9&32.0&23.7&14.3&82.6&33.8&80.7&49.6&8.2&84.3&28.3&33.1&0.0&6.9&0.2&39.2
\\
\hline
\end{tabular}}
  \caption{The ablation study about steps.}
  \label{tab:update}
\end{table*}
Tab.~\ref{tab:gamma} shows the detailed results about influence of $\gamma$.

\begin{figure*}[t]
    \centering
    \includegraphics[width=\linewidth]{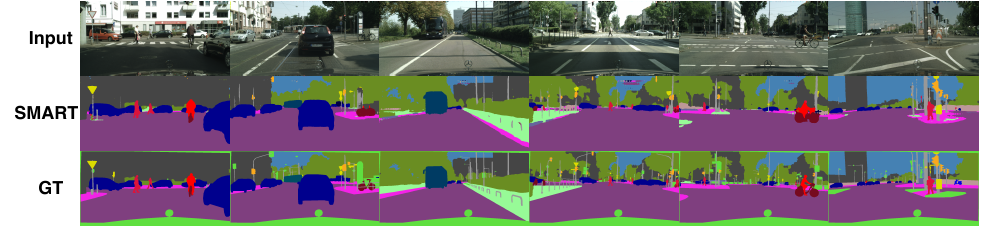}
    \caption{Qualitative results for GTA5-to-Cityscapes.}
    \label{fig:gta}
\end{figure*}

\begin{figure*}[t]
    \centering
    \includegraphics[width=\linewidth]{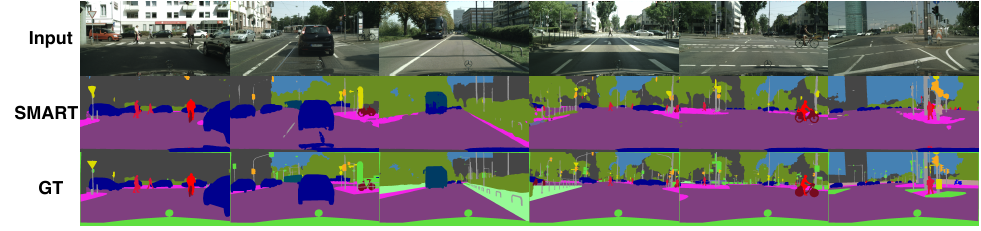}
    \caption{Qualitative results for SYNTHIA-to-Cityscapes.}
    \label{fig:syn}
\end{figure*}
\section{Ablation study}
\label{ablation}
In Tab.~\ref{tab:update}, to enable the semantic and depth features  more task-adaptive, we perform ablation studies about the steps of updating of function:

\begin{equation}
\begin{split}
\boldsymbol{\xi}_{n+1}&=\boldsymbol{\xi}_n-\gamma\left(\boldsymbol{\xi}_n-\boldsymbol{\nu} \operatorname{softmax}\left(\boldsymbol{\nu}^T \boldsymbol{\xi}_n\right)\right)\\
&=(1-\gamma)\boldsymbol{\xi}_n+\gamma\boldsymbol{\nu} \operatorname{softmax}\left(\boldsymbol{\nu}^T \boldsymbol{\xi}_n\right). \nonumber 
\end{split}
\end{equation}

We modify the steps in the fusion scheme from 0 to 4 and conduct several experiments to evaluate the performance of the proposed model. Our results show that: (1) When the step is 0, the performance is 53.4\%. Directly fusing semantic and depth features using fusion scheme 1 does not improve the performance compared with the baseline. This verifies that directly adding these features without iteration does not enhance the model's performance. (2) As the number of steps increases (up to 3), the corresponding performance gradually improves, demonstrating the effectiveness of iteratively extracting task-adaptive features using the Hopfield energy algorithm. Our model, SMART, achieves 57.1\% and 57.3\% mIoU with one- and two-step EB2F, respectively. Given the computational cost of the proposed model, we use one-step EB2F in the entire training process. (3) However, when the number of steps is more than 2, updating the depth features multiple times degrades the segmentation performance. This is because the repeated updating of depth features causes them to lose their geometric information, leading to a degradation in segmentation performance after feature fusion.
\section{Visualization}
\label{Visulization}
We provide a few qualitative examples on GTA5-to-Cityscapes and SYNTHIA-to-Cityscapes tasks in Fig.\ref{fig:gta} and Fig.\ref{fig:syn}, respectively.

% \section{TSNE}
% \label{TSNE}
% We also provide the TSNE \ref{van2008visualizing} visualization with more data samples in Fig.\ref{}, our SMART brings a significant improvement in distinguishing different categorical pixels in high-level feature space.

\section{Algorithm}
\label{algorithm}
The overall algorithm of SMART is shown in Algorithm.\ref{alg}.

\begin{algorithm}[t]
	\caption{The Proposed framework} 
	\label{alg} 
	\begin{algorithmic}[1]
	    \STATE \textbf{Input}: $\mathcal{D}_S=\left\{\left(\mathbf{x}_1^S, \mathbf{y}_1^S, \mathbf{d}_1^S\right), \ldots,\left(\mathbf{x}_n^S, \mathbf{y}_n^S, \mathbf{d}_n^S\right)\right\}$, $\mathcal{D}_T=\left\{\left(\mathbf{x}_1^T, \mathbf{d}_1^T\right), \ldots,\left(\mathbf{x}_m^T, \mathbf{d}_m^T\right)\right\}$; max iterations: $T_1$ and $T_2$
	    \\ \textbf{model}:  $f(\Theta)$;
	    \STATE  \textbf{Initialization}: Set $\Theta$;
	    \FOR{for t $\xleftarrow[]{}$ 1 to $T_1$}
    	    \STATE Compute the Hopfield energy between semantic and depth features:\\
                $E_{h}(\boldsymbol{\xi}; \boldsymbol{\nu})=\frac{1}{2} \boldsymbol{\xi}^T \boldsymbol{\xi}-\operatorname{lse}\left(\boldsymbol{\nu}^T \boldsymbol{\xi}\right)$;
                \STATE Obtain the task-adaptive features:\\
                $\boldsymbol{\xi}_{n+1}=(1-\gamma)\boldsymbol{\xi}_n+\gamma\boldsymbol{\nu} \operatorname{softmax}\left(\boldsymbol{\nu}^T \boldsymbol{\xi}_n\right)$;
                \STATE Obtain the fused semantic and depth features with two fusion schemes:\\
                $
                {\textit{Fusion}_{1}}(\boldsymbol{\xi},\boldsymbol{\nu}) = \boldsymbol{\xi}+\boldsymbol{\nu},
                $\\
                ${\textit{Fusion}_{2}}(\boldsymbol{\xi},\boldsymbol{\nu}) =  \boldsymbol{\nu} + (W_d^1 \otimes \boldsymbol{\xi})\odot \sigma(W_d^2\otimes\boldsymbol{\xi})$;
                 \STATE Compute the supervised losses for semantic segmentation and depth 
                estimation, respectively:\\
                $\mathcal{L}_{seg}^{total} = \mathcal{L}_{seg}^{S}+\mathcal{L}_{seg}^{T}+\tilde{\mathcal{L}}_{seg}^{S}+\tilde{\mathcal{L}}_{seg}^{T},$\\
                $\mathcal{L}_{dep}^{total} = \mathcal{L}_{dep}^{S}+\mathcal{L}_{dep}^{T}+\tilde{\mathcal{L}}_{dep}^{S}+\tilde{\mathcal{L}}_{dep}^{T},$\\
                $\mathcal{L}_{s}=\mathcal{L}_{seg}^{total}+\alpha\mathcal{L}_{dep}^{total};$
                \STATE Back propagation for $\mathcal{L}_{s}$;
         \STATE Update the model $\Theta$.
    	 \ENDFOR
      \FOR{for t $\xleftarrow[]{}$ 1 to $T_2$}
          \STATE Calculate the energy score for semantic segmentation and depth estimation:\\
          $E_{seg}(\mathbf{x})=-\log \sum_{i=1}^K e^{P^{i}(\mathbf{x})}$,\\
          $\tilde{E}_{seg}(\mathbf{x})=-\log \sum_{i=1}^K e^{\tilde{P}^{i}(\mathbf{x})}$,\\
          $E_{dep}(\mathbf{x})=\mathcal{L}_{dep},$\\
          $\tilde{E}_{dep}(\mathbf{x})=\tilde{\mathcal{L}}_{dep}$;\\
          \STATE Calculate the loss function of RFA:\\
          $\mathcal{L}_{\text{RFA}}= \mathcal{L}_{\text{RFA}}^{seg}+\alpha\mathcal{L}_{\text{RFA}}^{dep};$
          \STATE Obtain the overal objective:\\
         $ \mathcal{L}=\mathcal{L}_{s}+\beta\mathcal{L}_{\text{RFA}};$
          \STATE Back propagation for $\mathcal{L}$;
          \STATE Update the model $\Theta$.
      \ENDFOR
	    \STATE  \textbf{return}  $\Theta$.
	    \STATE  \textbf{End}.
	\end{algorithmic} 
\end{algorithm}

\section{Limitation and Future Work} 
\label{limitation}
In this paper, we mostly use the Hopfield energy function to build up our SMART framework by measuring the discrepancy between semantic and depth features. Future work will explore other energy functions to better learn task-adaptive features. Moreover, our SMART framework can be extended to explore the correlation between different tasks to improve the performance of these tasks, \eg, depth estimation.

%%%%%%%%%%%%%%%%%%%%%%%%%%%%%%%%%%%%%%%%%%%%%%%%%%%%%%%%%%%%%%%%%%%%%%%%%%%%%%%%
\clearpage
\bibliographystyle{./IEEEtran}
\bibliography{./IEEEabrv,./ref}
%% \bibliographystyle{./IEEEtran} % use IEEEtran.bst style
%% \bibliography{./IEEEabrv,./IEEEexample}